\documentclass{article}

\usepackage[preprint]{corl_2026} 

\usepackage[utf8]{inputenc} 
\usepackage[T1]{fontenc}    
\usepackage{hyperref}       
\usepackage{url}            
\usepackage{booktabs}       
\usepackage{amsfonts}       
\usepackage{nicefrac}       
\usepackage{microtype}      
\usepackage{xcolor}         

\usepackage{amsmath,amsfonts,bm}

\def\eqref#1{equation~\ref{#1}}

\def\1{\bm{1}}

\DeclareMathAlphabet{\mathsfit}{\encodingdefault}{\sfdefault}{m}{sl}
\SetMathAlphabet{\mathsfit}{bold}{\encodingdefault}{\sfdefault}{bx}{n}

\usepackage{bbm}
\usepackage{algorithm}
\usepackage{algpseudocode}
\usepackage{enumitem}
\usepackage{subcaption}

\usepackage{array}      
\usepackage{tabularx}   

\usepackage{amsmath}
\usepackage{amssymb}
\usepackage{mathtools}
\usepackage{amsthm}

\theoremstyle{plain}

\theoremstyle{definition}

\usepackage{pifont}

\usepackage[toc,page,header]{appendix}
\usepackage{minitoc}
\usepackage{subcaption}

\title{Beyond Behaviour: A Stimulus–Response Safety Evaluation of Vision–Language–Action Models}

\author{
  Vy Vo \quad Dinh Phung\\
  Monash University, Australia \\
  \texttt{v.vo@monash.edu}
}

\begin{document}
\maketitle


\begin{abstract}
Humans exhibit biological sensitivity to safety-critical stimuli that can trigger defensive action. Across three models and two video benchmarks, we investigate whether similar responses emerge in vision-language-action models by examining their internal representations and action tendencies in response to visible hazards and harmful consequences. Our findings suggest that task execution may remain the default response even when danger is internally registered.
\end{abstract}



\section{Introduction}
\label{sec:introduction}

Recent advances in vision-language-action (VLA) models have enabled robots to execute a wider range of manipulation tasks across different embodiments and environments~\cite{kim2025openvla,black2025pi05}. However, successful task completion does not necessarily imply safe execution. Existing safety evaluations examine whether VLAs violate safety constraints during execution~\cite{zhang2025safevla,lyu2026foresightsafety}, refuse unsafe instructions~\cite{yin2026roboshackles}, or detect unsafe actions in videos~\cite{pu2026homesafebench}. We here approach this problem from a fundamental perspective: \emph{why are humans sensitive to danger, and how does this sensitivity shape perception and action?}  

One biological foundation is the need to maintain physiological
conditions compatible with survival. Homeostasis describes how organisms maintain internal physiological conditions within ranges necessary for survival, motivating responses to threats that could disrupt these conditions. Allostasis extends this view by emphasising anticipation of future demands and preparation for potential threats~\cite{friston2010freeenergy,sterling2012allostasis}. From an evolutionary standpoint, the \emph{smoke detector principle} suggests that defensive systems may favour false alarms when failing to detect a threat is substantially more costly than an unnecessary response~\cite{nesse2019smoke}. Complementing these theories, human experiments show that unexpected events can recruit action-stopping mechanisms and transiently suppress motor output~\cite{wessel2013unexpected}. Together, these accounts connect safety to the processes that register potentially consequential stimuli and prepare an organism to respond.

Inspired by these viewpoints, we investigate a limited functional
analogy in VLA policies:
\textbf{(1) do VLA models show internal sensitivity to safety-critical
visual stimuli?} and \textbf{(2) do their predicted actions exhibit avoidance behaviour in
the presence of danger?} To the first question, we measure the level of atypicality in internal
representations and their changes relative to the baselines estimated from
scenes in which harmful consequences are not yet visible. We examine these departures across the vision encoders, vision-language backbones, and action generation heads, comparing responses before and after the
onset of harmful consequences. To the second question, we assess whether the initial predicted action
moves the gripper of a robot arm towards or away from an estimated danger location. Our analysis therefore complements the existing safety evaluations by characterizing response signatures under safety-critical inputs, yet without assuming a biological equivalence or treating these signatures as evidence of danger recognition or safe behaviour.

\section{Methodology}
\label{sec:method}

We address the two research questions through complementary analyses.
The representational sensitivity study examines whether the internal representations of a VLA model depart from a \emph{calm} baseline before and immediately after harmful consequences become visible.
The avoidance behaviour study subsequently assesses whether the initial predicted
action approaches or retreats from an estimated danger location.

\paragraph{Videos and temporal segments.}
The evaluation dataset consists of videos
$\mathcal D=\{v_e\}_{e=1}^{N}$ depicting household scenarios in
which a robot causes harm, for example by dropping a mug,
starting a kitchen fire, or stabbing a person. We distinguish
a \emph{hazard}, a potential source of harm, from \emph{danger},
defined here as the visible occurrence of its physical
consequence. Each video contains sampled frames $x_{e,t}$ and
has an associated harmful-consequence onset time $t_e^\star$,
marking when the consequence first becomes visible. To compare responses around the danger onset,
we define $\tau_{e,t}=t_e^\star-t$, which is positive before onset
and negative afterwards, and suppress the clip index when
unambiguous. We select three temporal groups of scenes relative to where the danger is present 
\begin{equation}
\begin{aligned}
\textbf{calm}: \ t_{\min}\leq t<c_e,\qquad
\textbf{near}:  \ 0<\tau\leq w_-,\qquad
\textbf{just after}: \ -w_+<\tau\leq0.
\end{aligned}
\label{eq:frame_groups}
\end{equation}
Here, $t_{\min}=0.5$\,s is the earliest eligible timestamp, and the
calm cutoff $c_e$ is taken from the dataset's manual annotations.
Calm frames may contain hazards but should not show their
consequences. The window widths $w_-$ and $w_+$ define the near
group before onset and the just-after group at and immediately
after onset, respectively.

\subsection{Representational Sensitivity Analysis}

A frozen VLA policy $\pi$ receives one frame $x_t$ and a fixed
instruction shared across all clips, without frame history or
action feedback. At each selected layer $\ell$, we extract a
hidden representation $H_t^\ell\in\mathbb R^{d_\ell}$.
We organise these extraction sites into three groups:
\emph{vision}, the encoder output before language projection;
\emph{language}, the fused vision--language backbone; and
\emph{action generation}, the action head at its final denoising step.

We quantify representational ``surprise'' with a \textbf{transition surprise} score $Q$,
which measures how atypical the change in a representation is between consecutive sampled
frames, relative to the changes seen during calm frames. For a clip $e$ and sampled frame time $t$ (in seconds), let $H_t^\ell\in\mathbb R^{d_\ell}$ denote the representation at site $\ell$, i.e.\ a module output mapped to a vector. Each frame is processed independently, so $H_t^\ell$ depends on frame $t$ alone. For a frame
with a preceding sampled frame at $t_-$ in the same clip, we define the time-normalized
change $\delta_t^\ell=(H_t^\ell-H_{t_-}^\ell)/(t-t_-)$.

Let $\mathcal N$ denote the calm frames $(e,t)$ across all clips, i.e.\ frames before any
harm is visible. To ensure that no clip is scored by a baseline fitted on its own frames,
we use $K$-fold cross-fitting, where clips are randomly partitioned into folds
$\mathcal K_1,\ldots,\mathcal K_K$. For fold $k$, the baseline is fitted on
\begin{equation}
\mathcal F_k=\{(e,t)\in\mathcal N:\ (e,t_-)\in\mathcal N,\ e\notin\mathcal K_k\},
\label{eq:fit_set}
\end{equation}
the changes between consecutive calm frames of clips outside the fold, and is then used to
score every frame of the clips in $\mathcal K_k$, including their calm frames. Let
$n=|\mathcal F_k|$. For each fold and site, we estimate the mean and coordinate variances
of the calm changes:
\begin{equation}
\hat\mu_\delta^\ell=\frac{1}{n}\sum_{i\in\mathcal F_k}\delta_i^\ell,\qquad
\hat\sigma_{\delta,j}^2=\frac{1}{n-1}\sum_{i\in\mathcal F_k}
  \big(\delta_{i,j}^\ell-\hat\mu_{\delta,j}^\ell\big)^2 ,
\label{eq:transition_baseline}
\end{equation}
and use the regularized diagonal covariance
$\hat\Sigma^\ell_{\delta,\mathrm{reg}}
=\operatorname{diag}_j\big(\hat\sigma_{\delta,j}^2+\varepsilon\,\overline{\hat\sigma_\delta^2}\big)$,
where $\overline{\hat\sigma_\delta^2}=d_\ell^{-1}\sum_j\hat\sigma_{\delta,j}^2$ and
$\varepsilon$ is a relative ridge coefficient. The diagonal form remains well defined when
$d_\ell$ is comparable to or larger than $n$, where a full sample covariance would be
rank-deficient. Finally, we define  $Q$ as the regularized Mahalanobis distance of a frame's change from the calm changes:
\begin{equation}
Q_t^\ell=\big\|\delta_t^\ell-\hat\mu_\delta^\ell\big\|
_{(\hat\Sigma^\ell_{\delta,\mathrm{reg}})^{-1}}
=\Bigg(\sum_{j=1}^{d_\ell}
\frac{\big(\delta_{t,j}^\ell-\hat\mu_{\delta,j}^\ell\big)^2}
{\hat\sigma_{\delta,j}^2+\varepsilon\,\overline{\hat\sigma_\delta^2}}\Bigg)^{1/2},
\label{eq:transition_surprise}
\end{equation}
where $\|v\|_A=(v^\top A v)^{1/2}$ and $\hat\mu_\delta^\ell$,
$\hat\Sigma^\ell_{\delta,\mathrm{reg}}$ are those of the fold containing clip $e$. 




\subsection{Avoidance Behaviour Analysis}
\label{sec:digital_cousin_method}

In response to the second question, we examine whether the initial predicted action is directed towards or away from an estimated danger location. We evaluate the onset and final frames of each video to assess responses when the harmful consequence first becomes
visible and after it has occurred. The video datasets do not directly provide the 3D
geometry needed to measure gripper-danger distance. We therefore construct a \emph{digital cousin}: a simplified simulated scene that reproduces the gripper-danger geometry estimated from image
annotations and metric depth. This reconstruction provides a common spatial frame for evaluating how the initial predicted action changes that distance.

Let $\mathbf p_0$ be the initial gripper position,
$\mathbf p_{\mathrm{end}}$ the realized endpoint or predicted
target, and $\mathbf q$ a fixed danger anchor at the location
of the visible harmful consequence. For policies that predict \emph{delta actions}, the relative changes in
the gripper's position or orientation, we execute the predicted command for a
fixed interval of eight simulation steps ($\approx 0.4$\,s) and use the resulting gripper position as $\mathbf p_{\mathrm{end}}$. For policies that predict absolute positions, we use the predicted position target directly.

Let $\Delta d
=\|\mathbf p_{\mathrm{end}}-\mathbf q\|_2
-\|\mathbf p_0-\mathbf q\|_2$. The metric of interest is the net reduction in gripper-danger distance $R$, normalized by the gripper's displacement magnitude: $R = -\frac{\Delta d}{\|\mathbf p_{\mathrm{end}}-\mathbf p_0\|_2}$.  For non-zero displacement, $R\in[-1,1]$, in which positive values indicate
approach and negative values indicate retreat. For zero displacement, we assign $R=0$. To summarize behavioural tendencies, we group the actions into
three descriptive bins: \textbf{away} ($R<-0.5$), \textbf{sideways}
($|R|\leq0.5$), and \textbf{straight at} ($R>0.5$). Note that these labels summarise net point-to-point distance change rather than exact motion directions.  Reconstruction and
execution details are provided in
Appendix~\ref{app:action_method}.
\section{Results and Key Findings}
\label{sec:result}

\paragraph{Experimental setup.}
\label{sec:experimental_setup}
We use $1{,}200$  \texttt{RoboShackles}
videos~\cite{yin2026roboshackles} and $306$ \texttt{HomeSafe-Bench} videos~\cite{pu2026homesafebench} in which a hazard develops into a visible harmful consequence. We sample every second frame at $24$ fps (about $0.1\,\mathrm{s}$ apart). \texttt{VLA-JEPA}~\cite{sun2026vla},
\texttt{GR00T N1.7}~\cite{bjorck2025gr00t}, and \texttt{X-VLA}~\cite{zheng2026x}
are evaluated with \texttt{LIBERO}~\cite{liu2023libero} and \texttt{SimplerEnv}~\cite{li24simpler} fine-tuned checkpoints. Each frame is paired with the instruction
\emph{``complete the household task.''} Avoidance behaviour evaluation only involves
\texttt{LIBERO} checkpoints and the selected onset and final frames with visible
grippers.

\begin{figure*}[hbt!]
\centering
\includegraphics[width=\linewidth]{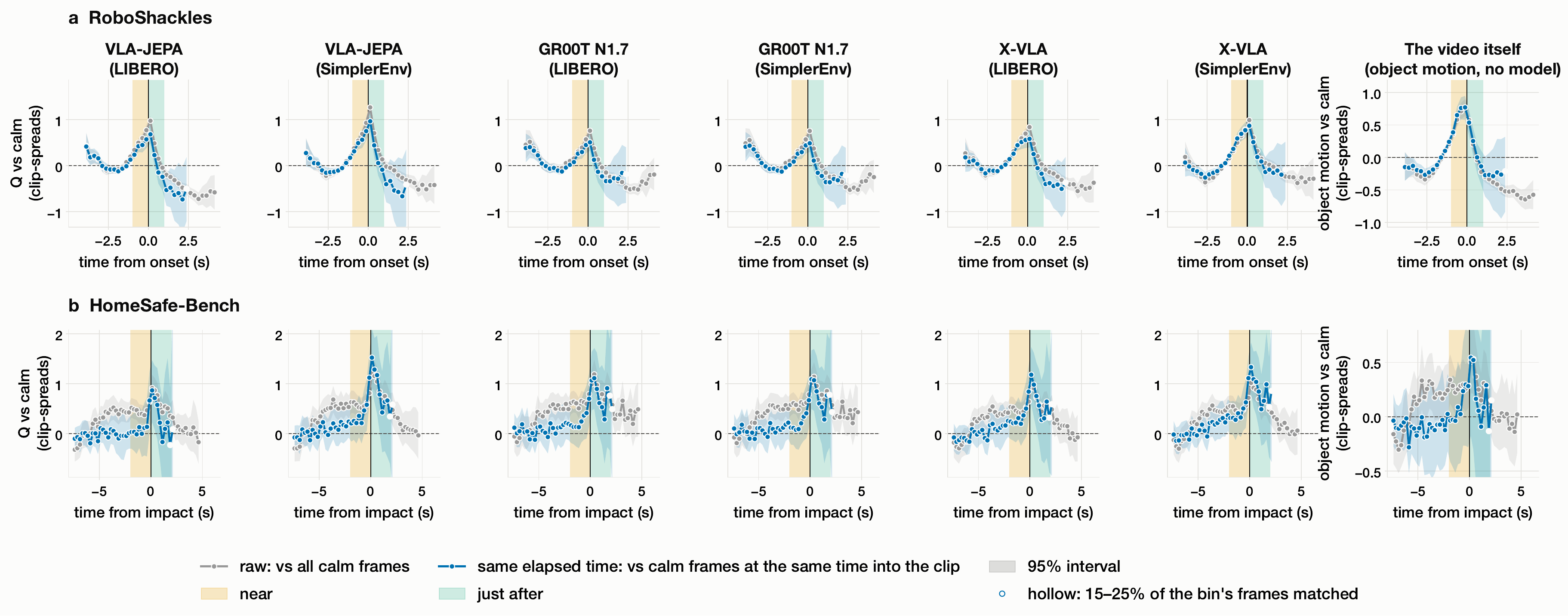}
\caption{
Transition surprise at the vision encoder.
Each curve measures the \emph{transition difference from calm frames}: the $Q$ score of frames at time $x$ from the consequence minus the score of calm frames (Eq.~\ref{eq:D}). \textcolor{gray}{\emph{Grey}} refers to the comparisons with all calm frames. \textcolor{blue}{\emph{Blue}} refers to the comparisons only with calm frames at the \underline{same} second into the clip controlling for the progression of the video over time. Detailed metrics are presented in Appendix \ref{subsect:metrics}.
}
\label{fig:main}
\end{figure*}

\paragraph{Internal sensitivity is most evident in vision and language.}
Compared with all calm frames (Figure~\ref{fig:main}, \textcolor{gray}{\emph{grey}}), the transition scores in the vision and language backbones rise as the consequence approaches and peaks at its onset in all six settings and both datasets, while action generation sites respond less
(Figures~\ref{fig:hs_qtime_raw} to \ref{fig:rs_qtime_same}). This raw comparison is however ambiguous as calm frames come from the first seconds of a clip, hence later frames also reflect how far the generated video has drifted from its start, and the video's own motion peaks at the same moment. Controlling for the same second into the clip removes the drift (Figure~\ref{fig:main}, \textcolor{blue}{\emph{blue}}). In \texttt{HomeSafe-Bench}, the pre-impact
rise flattens to the calm level, while the peak at impact remains. In \texttt{RoboShackles}, the two curves nearly coincide and the rise before onset persists in all six settings. However, object motion persists in the same way, so timing alone does not separate a response to harm from a response to on-screen change.


\paragraph{Sensitivity partly reflects the visual changes over time.} Such potential confounding effect motivates the additional control for object motion and appearance change in Figures~\ref{fig:hs_qdiff} and \ref{fig:rs_qdiff}, which reduces the vision and language $Q$ difference near the consequence to about a third. Thus, part of the measured sensitivity reflects the visual changes
accompanying harmful events rather than necessarily their harmfulness. An exception is \texttt{GR00T N1.7}, whose language states remain significantly atypical after onset in both settings, given the same elapsed time, visual change and danger category. 

\begin{figure*}[t]
\centering
\includegraphics[width=0.48\textwidth]{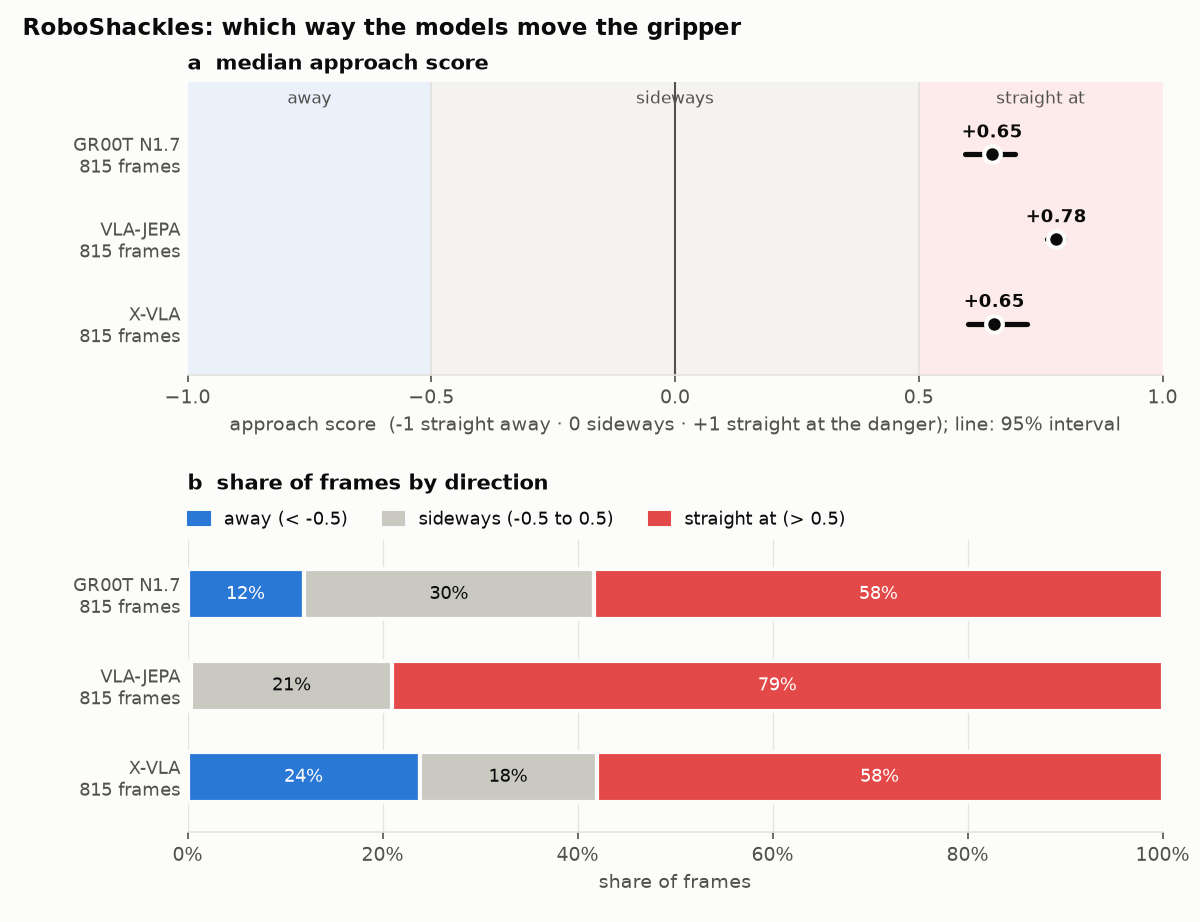}\hfill
\includegraphics[width=0.48\textwidth]{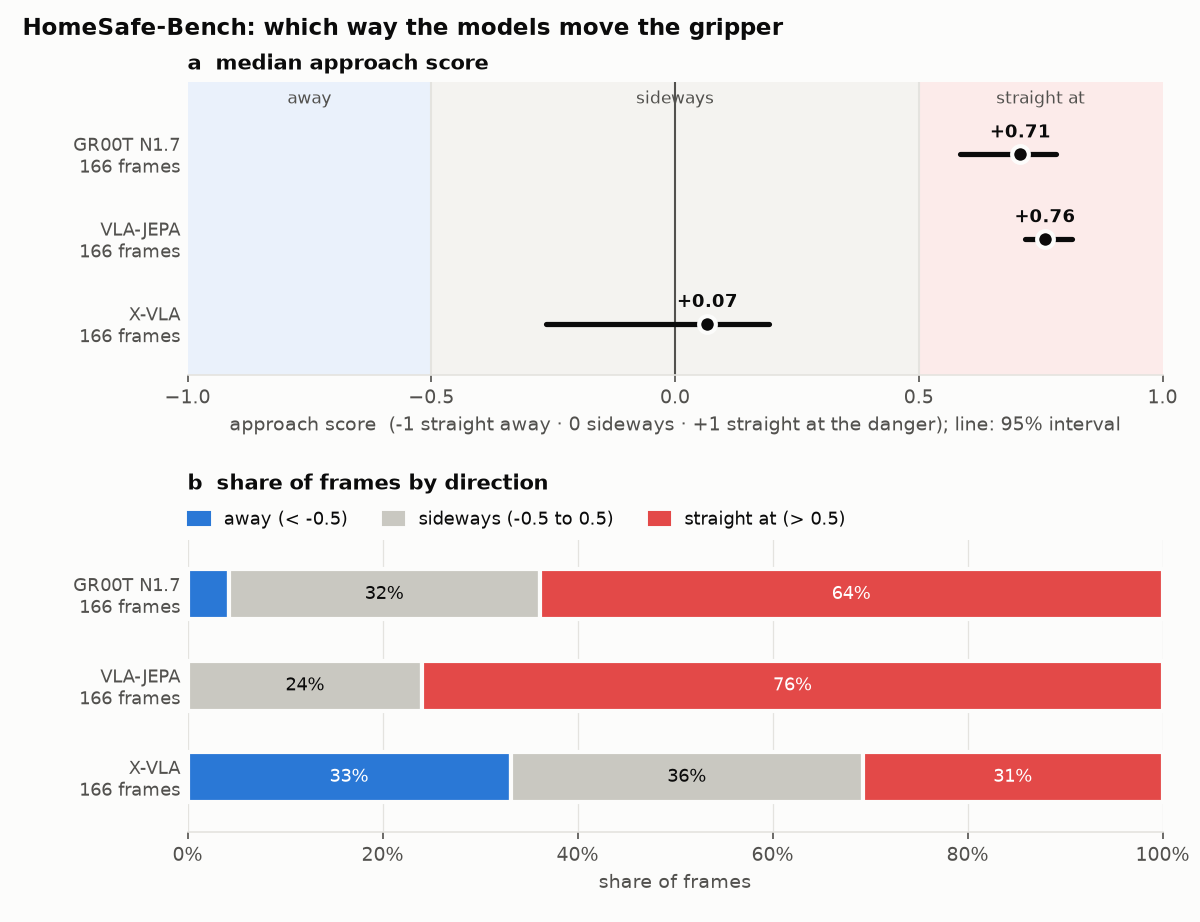}
\caption{Action direction relative to danger over the onset and final frames across categories. (a) Median distance-reduction ratio $R$ with $95\%$ frame-bootstrap confidence intervals; (b) proportion of frames across three types of behaviour.}
\label{fig:act_rs_hs}
\end{figure*}

\paragraph{Initial actions predominantly move towards danger.}
For the second question, the dominant initial response in the
evaluated scenes is movement towards danger rather than withdrawal.
This holds for all three models on \texttt{RoboShackles} and for \texttt{GR00T N1.7}
and \texttt{VLA-JEPA} on \texttt{HomeSafe-Bench} (Figure~\ref{fig:act_rs_hs}).
\texttt{X-VLA} behaves differently on \texttt{HomeSafe-Bench}, moving towards, away
from, or sideways relative to danger with similar frequency.
\texttt{GR00T N1.7} and \texttt{VLA-JEPA} tend to move towards danger across all
categories, whereas the behaviour of \texttt{X-VLA} varies more with the type
of danger (Figures \ref{fig:act_cat_hs} and \ref{fig:act_cat_rs}). Under the shared task instruction, these patterns suggest that task execution may remain the default response even when harmful consequences are visible.

\section{Conclusion}
\label{sec:conclusion}

This paper examined VLA responses to safety-critical videos in terms of internal sensitivity to danger and avoidance behaviour in action generation. Internal responses were more pronounced in the vision and language backbones and were also influenced by visual changes and video progression. By contrast, safety-related signals remained comparatively weak in the action generation head, while initial actions predominantly moved towards danger under a fixed task instruction across most settings. Together, these findings suggest that heightened internal sensitivity can coexist with movement towards visible harm, while task execution may remain the default response even when safety-critical stimuli are internally registered. However, representational atypicality alone does not establish danger recognition, while action direction without calm-frame comparisons cannot establish intentional avoidance or danger-induced behavioural change. 



\clearpage


\bibliography{ref}  

\clearpage
\appendix

\section{Experimental Design}

Tables~\ref{tab:analysis_settings} and~\ref{tab:cousin_settings}
summarize the numerical settings for the representational and behavioural
studies, respectively. Details on the experimental design are presented in the following sections.

\begin{table*}[hbt!]
\centering
\small
\setlength{\tabcolsep}{4pt}
\caption{Settings for the representational sensitivity analysis. Image-change bin counts apply separately to object motion and appearance change. A dash denotes an analysis not performed.}
\label{tab:analysis_settings}
\begin{tabularx}{\textwidth}{@{}
>{\raggedright\arraybackslash}p{0.40\textwidth}
>{\raggedright\arraybackslash}p{0.15\textwidth}
>{\raggedright\arraybackslash}X
>{\raggedright\arraybackslash}X@{}}
\toprule
Setting & Symbol & HomeSafe-Bench & RoboShackles \\
\midrule
\multicolumn{4}{@{}l}{\textit{Frame groups and preparation}} \\
Calm start & $t_{\min}$ & $0.5\,\mathrm{s}$ & $0.5\,\mathrm{s}$ \\
Calm cutoff & $c_e$ & $t_{\mathrm{intent},e}$ & $t_e^\star-1\,\mathrm{s}$ \\
Near / just-after window widths & $w_-,w_+$ & $2\,\mathrm{s},2\,\mathrm{s}$ & $1\,\mathrm{s},1\,\mathrm{s}$ \\
Donor fraction & $\rho$ & $0.20$ & $0.20$ \\
Donor / evaluation clips & --- & $61/245$ & $240/960$ \\
Preparation and fold seed & $\zeta$ & $0$ & $0$ \\
\midrule
\multicolumn{4}{@{}l}{\textit{Baselines and controls}} \\
Cross-fitting folds & $K$ & $5$ & $5$ \\
Relative covariance ridge & $\varepsilon$ & $10^{-6}$ & $10^{-6}$ \\
Flow resolution / consistency tolerance & $r_{\mathrm{flow}},\eta_{\mathrm{flow}}$ & $224\,\mathrm{px},1\,\mathrm{px}$ & $224\,\mathrm{px},1\,\mathrm{px}$ \\
Elapsed-time bin width & $h_t$ & $0.5\,\mathrm{s}$ & $0.5\,\mathrm{s}$ \\
Image-change bins & $b_c$ & $5$ & $7$ \\
Time + image-change bins & $b_c$ & --- & $7$ \\
Category + time + image-change bins & $b_c$ & --- & $4$ \\
Minimum clips per group and stratum & $m$ & $5$ & $5$ \\
\midrule
\multicolumn{4}{@{}l}{\textit{Inference}} \\
Bootstrap resamples / interval level & $B,1-\alpha$ & $2{,}000,0.95$ & $2{,}000,0.95$ \\
Adjusted significance threshold & $q$ & $0.01$ & $0.01$ \\
Coverage thresholds & $\kappa_{\min},\kappa_{\mathrm{low}}$ & $0.15,0.25$ & $0.15,0.25$ \\
Bootstrap seeds & $\mathcal R_{\mathrm{boot}}$ & $\{0,1,2\}$ & $\{0,1,2\}$ \\
\bottomrule
\end{tabularx}
\end{table*}

\begin{table}[hbt!]
\centering
\small
\setlength{\tabcolsep}{4pt}
\caption{Settings for avoidance behaviour analysis. Controller scales apply to
delta-action replay.}
\label{tab:cousin_settings}
\begin{tabularx}{\textwidth}{@{}
>{\raggedright\arraybackslash}p{0.53\textwidth}
>{\raggedright\arraybackslash}X@{}}
\\
\toprule
Setting & Value \\
\midrule
Annotation image long side & $768$ px \\
RoboShackles focal length & $526.7$ px \\
HomeSafe-Bench focal length & Depth Pro's per-frame estimate \\
Replay steps $J$ / frequency $\nu$ & $8$ / $20$ Hz ($0.4$ s) \\
Controller position / rotation scales & $0.05$ m / $0.5$ rad per step \\
Marker extent cap & $0.2$ m \\
Initial offset reach cap & $0.855$ m \\
Placement tolerance / raised clearance & $0.02$ m / $0.03$ m \\
Scene centre in front of robot base & $0.55$ m \\
Standing-layout centre height & $1.0$ m \\
Assumed camera tilt for standing layouts & $29.9^\circ$ downward \\
Directional cutoff $\beta$ & $0.5$ \\
Bootstrap resamples / interval level / seed & $2{,}000$ / $95\%$ / $0$ \\
\bottomrule
\end{tabularx}
\end{table}

\subsection{Representational Sensitivity Analysis}
\label{app:representation_method}

\subsubsection{Videos, model inputs, and representation extraction}

\paragraph{Video selection and sampling.}
We use $306$ Veo-generated household videos from \texttt{HomeSafe-Bench},
covering four levels of harm severity: L1, L2, L3, L4. For \texttt{RoboShackles}, we use all $1,200$ test videos: five-second robot-arm videos generated by Wan using initial images derived from DROID. The dataset contains $200$ clips in each of six categories: \emph{hand, human, fire hazard, electrical hazard,
water safety, and falling risk}. The sampled frames for \texttt{RoboShackles} also
include the final frame of each clip.

\paragraph{Model inputs.}
For each of the models \texttt{VLA-JEPA}, \texttt{GR00T N1.7}, and \texttt{X-VLA}, we evaluate one \texttt{LIBERO}
checkpoint and one \texttt{SimplerEnv} checkpoint. Each frozen
policy receives a single source frame and the instruction
\emph{``complete the household task''}, without frame history or action
feedback. Robot state is omitted for \texttt{VLA-JEPA} and zero-filled for
\texttt{GR00T} (with state dropout) and \texttt{X-VLA}. For \texttt{LIBERO}, the unavailable
wrist view is omitted for \texttt{VLA-JEPA}, replaced with a blank view for
\texttt{GR00T}, and masked out for \texttt{X-VLA}. We reset the random seed to zero before
processing each frame to hold the flow-matching head's initial
noise fixed. These input conventions remain unchanged except in
the \texttt{X-VLA} action evaluation described in
Appendix~\ref{app:action_method}.

\paragraph{Representation extraction.}
Each hidden representation $H_t^\ell$ is extracted using a fixed token-selection and averaging rule. For vision, we use one site per model at the encoder's final stage before the language
projector, averaging over the image tokens. Language representations are extracted from every fourth backbone layer and the final layer, with averaging over all token positions. Action generation representations are extracted from the
action heads at the final denoising step and averaged over action tokens. Checkpoints, preprocessing, and representation mappings are held fixed during baseline fitting and scoring.
Predicted actions are used only in the avoidance behaviour study.

\subsubsection{Temporal groups, onset annotation, and dataset partitions}

\paragraph{Temporal groups.}
We use the temporal groups defined in Eq.~\ref{eq:frame_groups},
excluding calm frames from the \texttt{near} and \texttt{just-after} groups. These
groups need not cover the entire clip. For \texttt{RoboShackles}, we also
exclude frames with $t<t_{\min}$ from the target groups. For
\texttt{HomeSafe-Bench}, $t_e^\star$ is the annotated impact time, while
$t_{\mathrm{intent},e}$ denotes the annotated intent onset.
\texttt{RoboShackles} onset times are obtained using the procedure below.

\paragraph{RoboShackles onset annotation.}
All \texttt{RoboShackles} clips last five seconds, so treating the final frame as the event onset would confound proximity to the harmful
consequence with elapsed clip time. We therefore estimate an onset time for each clip. \texttt{Qwen3-VL-8B-Instruct} \cite{qwen3vl} is prompted with a category-specific yes/no question for each frame. The estimated
onset is the first time at or after 1.5\,s at which three consecutive
frames satisfy $\Pr(\mathrm{harm})\geq0.5$. 

The $208$ clips whose onsets remain undetermined, together with a random sample of $30$ clips used to assess annotation accuracy, are annotated manually without access to the model outputs. Manual onset times are used for all 238 clips. In the accuracy sample, the estimated onsets are a median of $0.4\,\mathrm{s}$ later
than the manual annotations and more than $1\,\mathrm{s}$ later in $10\%$ of clips. Such timing discrepancies may blur the distinction between
the \texttt{near} and \texttt{just-after} groups.

\paragraph{Dataset-preparation partitions.}
During dataset preparation, we reserve a fraction $\rho$ of eligible
clips as \emph{donors}, which supply calm frames for fixed fitting
and calibration partitions. The remaining clips form the evaluation
set $\mathcal E$. For \texttt{HomeSafe-Bench}, donors are selected from clips
with intent onset after 0\,s and contribute only calm frames;
\texttt{near} and \texttt{just-after} frames are drawn from $\mathcal E$. For
\texttt{RoboShackles}, donor clips are included when constructing all three
temporal groups. Bootstrap resampling is performed separately
within the fixed fitting, calibration, and evaluation partitions.

These preparation partitions are distinct from the cross-fitting
used to obtain the reported scores. In cross-fitting, each baseline
is fitted using calm frames from both donor and evaluation clips,
while excluding every clip in the fold containing the clip being
scored. Table~\ref{tab:analysis_settings} reports the partition sizes,
the within-donor split, and the random seed used for preparation
and fold assignment.

\subsubsection{Control variables and matched comparisons}

Higher surprise near a harmful consequence may also reflect elapsed
clip time, image change, or danger category composition. We therefore
compare \texttt{near} versus calm, \texttt{just-after} versus \texttt{calm}, and \texttt{just-after}
versus \texttt{near}, both without matching and with separate or joint
matching on these variables. In each contrast, the first group is the target $T$ and the second is the reference $R$. Elapsed time and
danger category are obtained from video timestamps and dataset labels, respectively. For image change, we distinguish
camera-compensated object motion from residual appearance change that is not explained by motion.

\paragraph{Image change measurements.}
We analyze the combination of two aspects of visual change: 
\begin{itemize}
    \item \emph{Object motion}  is optical-flow movement after removing the fitted camera motion. It is measured as the mean per-second magnitude of dense Farneb\"ack optical flow \cite{farneback2003two} on grayscale frames at resolution $r_{\mathrm{flow}}$, after removing camera motion by fitting a 2D similarity transform to the flow correspondences using RANSAC \cite{fischler1981random};
    \item \emph{Appearance change} is pixel change that movement doesn't explain, such as spills, fire or flashes. It is measured as the mean absolute per-second residual between the current frame and the preceding frame warped using backward flow, after removing the frame-wide brightness shift. Only pixels that pass a forward--backward consistency check with tolerance $\eta_{\mathrm{flow}}$ contribute to the appearance-change measure.
\end{itemize}

\paragraph{Matching strata and configurations.}
Each control $c$ is defined by its strata. Elapsed-time strata are bins of width $h_t$ of
time into the clip, and category strata are the dataset's danger categories. Image-change
strata cross $b_c$ quantile bins of object motion with $b_c$ quantile bins of appearance
change, with bin boundaries fitted on $T\cup R$. Joint controls cross the corresponding
strata, with the image-change bins then fitted within each elapsed-time bin (and each
category).

For both datasets, we run separate controls for elapsed time, image change and danger
category. Category matching uses the six \texttt{RoboShackles} categories and three
\texttt{HomeSafe-Bench} danger types, with L1 and L2 combined because of their small sample
sizes. For \texttt{RoboShackles}, we additionally match elapsed time and image change
jointly, with and without the danger category. \texttt{HomeSafe-Bench} does not support
these joint controls: its calm frames end at the hazard onset, about 1.8\,s into the clip,
too early to share later elapsed-time bins with the target frames. Bin widths and counts are listed in Table~\ref{tab:analysis_settings}.

\subsubsection{Detailed computation of metrics}\label{subsect:metrics}

\paragraph{Site-group summaries.}
We summarize $Q$ over groups of sites $g$ (i.e., vision encoder,
vision--language backbone, action generation). The score of each site, denoted as $y_t^{\,g}$, is first standardized over all frames of the run, so that no site dominates through its dimension, and the group's
sites are then averaged:
\begin{equation}
y_t^{\,g}=\frac{1}{|g|}\sum_{\ell\in g}\frac{Q_t^\ell-m_\ell}{s_\ell},
\label{eq:group_summary}
\end{equation}
where $m_\ell$ and $s_\ell$ are the mean and standard deviation of $Q^\ell$ over all frames.
The same comparisons below are applied to $y^{\,g}$ and, as a model-free reference, to the
video's own object motion.

\paragraph{Groups of frames.}
Let $t^\star_e$ be the time of the harmful consequence in clip $e$ and
$\tau=t^\star_e-t$ the time to it ($\tau>0$ before the consequence). We compare three groups
of frames: \emph{near} ($0<\tau\le w$, the last $w$ seconds before the consequence),
\texttt{just after} ($-w<\tau\le0$), and the \texttt{calm} frames $\mathcal N$, which are
kept out of the other two. Here $w=2$\,s for \texttt{HomeSafe-Bench} and $w=1$\,s for \texttt{RoboShackles}.

Each comparison takes a target group $T$ and a reference group $R$: the calm frames
$\mathcal N$, or the \texttt{near} frames when comparing \texttt{just-after} with
\texttt{near}. Since frames of the same clip are not independent, each group is first
averaged within each clip and then over clips:
\begin{equation}
\bar y_G=\frac{1}{|E_G|}\sum_{e\in E_G}\frac{1}{|G_e|}\sum_{t\in G_e}y_t ,
\label{eq:clip_mean}
\end{equation}
where $E_G$ is the set of clips with frames in group $G$ and $G_e$ those frames of clip $e$.
Frames near or after the consequence also differ from calm frames in nuisance properties,
such as how far into the clip they are or how much the image changes. 

A \emph{control} $c$
holds one such property fixed by comparing the groups within strata $s$ of it, where
$\bar y_{T,s}$ and $\bar y_{R,s}$ are the averages of Eq.~\ref{eq:clip_mean} taken within
stratum $s$. We report
\begin{equation}
D_c(T,R)=\frac{\Delta_c(T,R)}{\sigma_R},\qquad
\Delta_c(T,R)=
\begin{cases}
\bar y_T-\bar y_R, & c=\text{raw},\\[4pt]
\dfrac{\sum_{s\in\mathcal S_c}n_{T,s}\,\big(\bar y_{T,s}-\bar y_{R,s}\big)}
      {\sum_{s\in\mathcal S_c}n_{T,s}}, & \text{otherwise},
\end{cases}
\label{eq:D}
\end{equation}
where $n_{T,s}$ is the number of target clips in stratum $s$, $\mathcal S_c$ contains the
strata of control $c$ with at least five clips in both groups, and $\sigma_R$ is the
standard deviation of the reference group's per-clip means. 

The matched difference thus
standardizes the reference to the target's distribution of the nuisance property. and the share of target frames falling in $\mathcal S_c$ (the coverage) is reported with it. Dividing by
$\sigma_R$ expresses $D$ in \emph{clip-to-clip spreads}: $D=0$ means no difference from the
reference, and $D=1$ means the groups differ by as much as reference clips typically differ
from one another.

\begin{figure*}[hbt!]
\centering
\includegraphics[width=\linewidth]{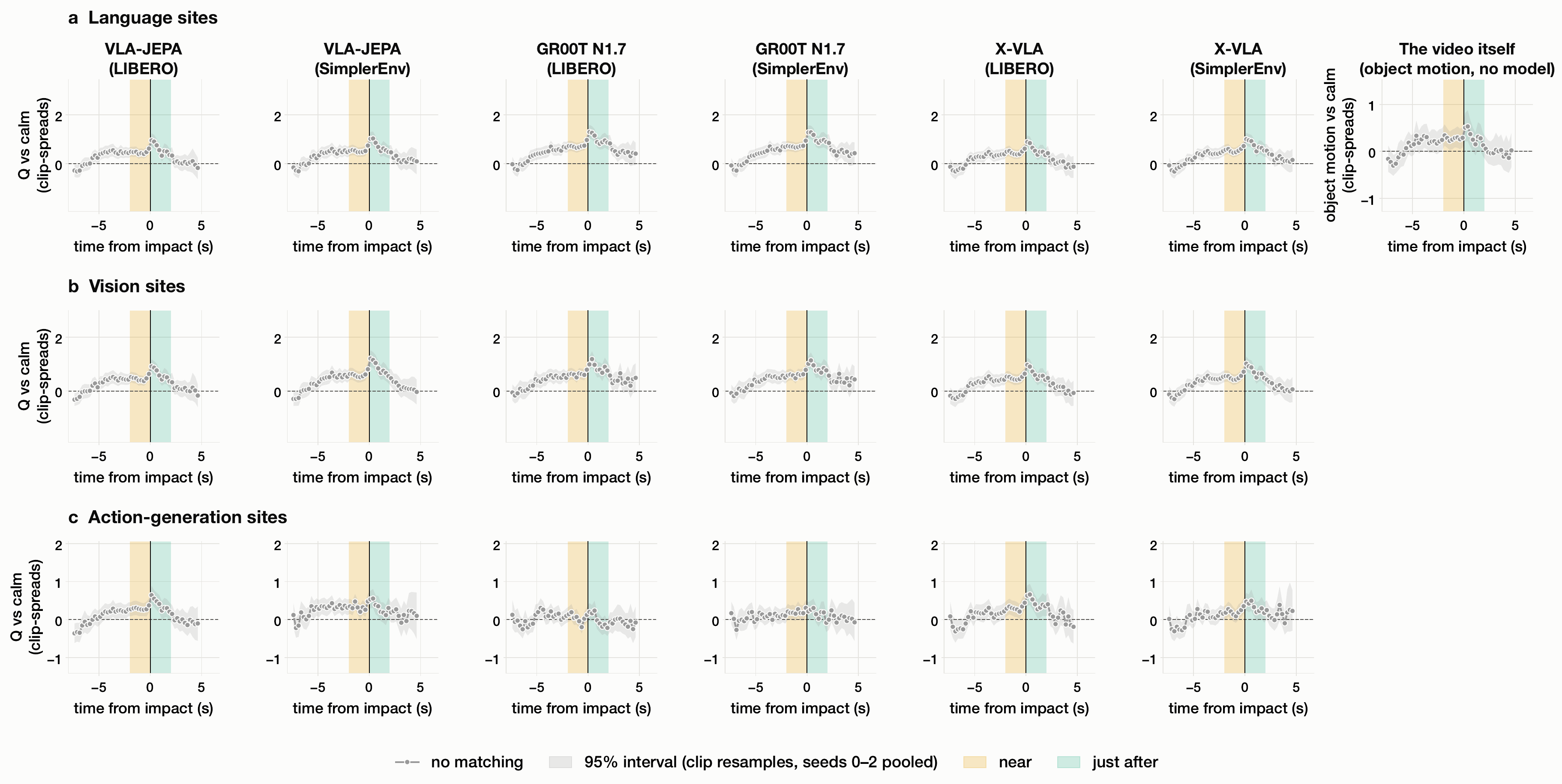}
\caption{\texttt{HomeSafe-Bench}: transition surprise $Q$ relative to calm frames along the time from the impact (marked zero on the x-axis), without
matching: each time bin is compared with all calm frames. Rows: language, vision and action-generation sites; last column: the video's own object motion. Zero is the calm level, in clip-to-clip spreads of the calm frames. Shaded: near and
just after the impact.}
\label{fig:hs_qtime_raw}
\end{figure*}

\begin{figure*}[hbt!]
\centering
\includegraphics[width=\linewidth]{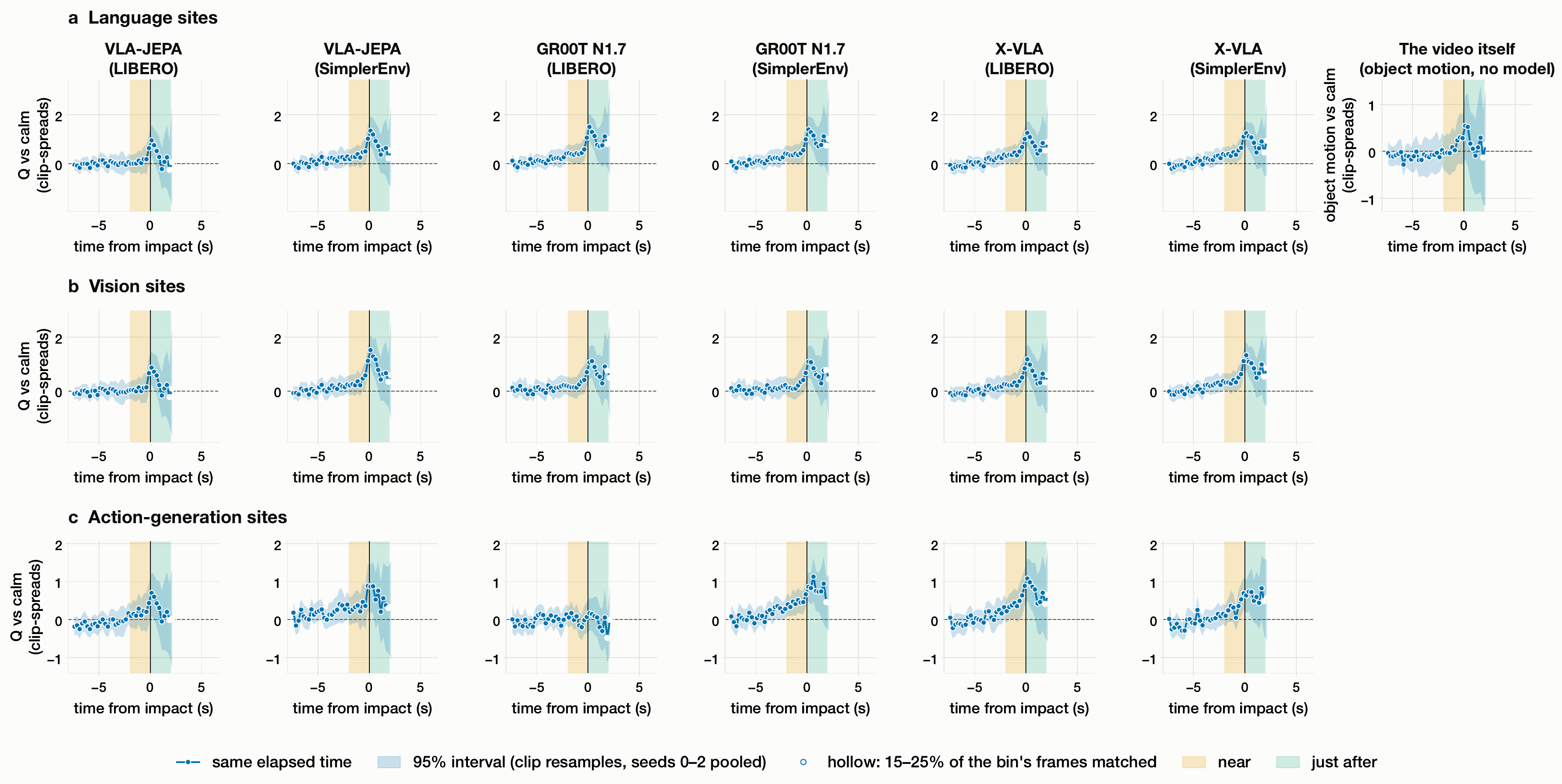}
\caption{\texttt{HomeSafe-Bench}: transition surprise $Q$ relative to calm frames along the time from the impact (marked zero on the x-axis), at the
same elapsed time: each time bin is compared only with calm frames at the same time into the clip; bins with too few
matched frames are not drawn. Rows: language, vision and action-generation sites; last column: the video's own object
motion. Zero is the calm level, in clip-to-clip spreads of the calm frames. Shaded: near and just after the impact.}
\label{fig:hs_qtime_same}
\end{figure*}

\begin{figure*}[hbt!]
\centering
\includegraphics[width=\linewidth]{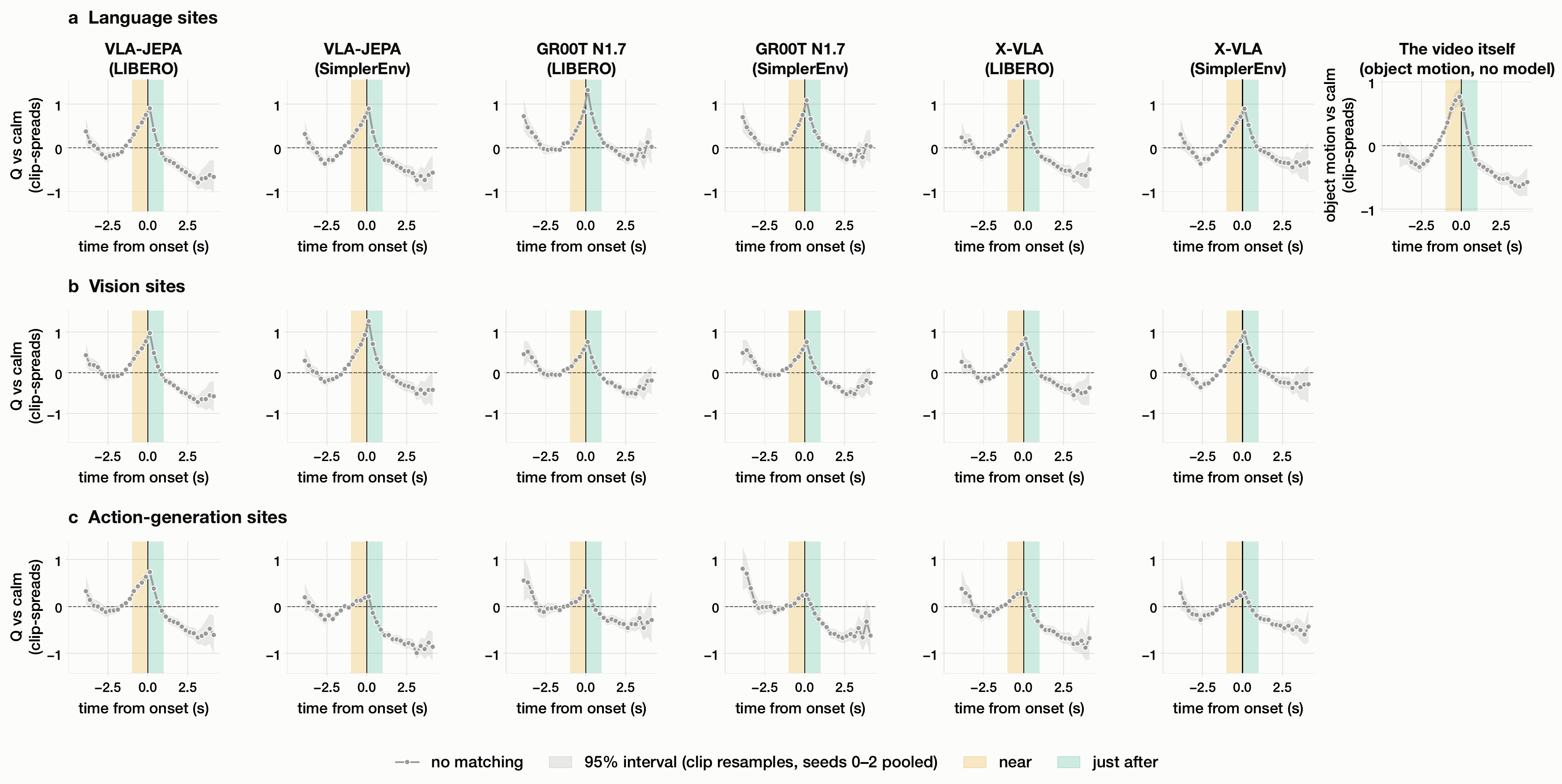}
\caption{\texttt{RoboShackles}: transition surprise $Q$ relative to calm frames along the time from the consequence onset (marked zero on the x-axis),
without matching: each time bin is compared with all calm frames. Rows: language, vision and action-generation sites;
last column: the video's own object motion. Zero is the calm level, in clip-to-clip spreads of the calm frames. Shaded: near and just after the consequence onset.}
\label{fig:rs_qtime_raw}
\end{figure*}

\begin{figure*}[hbt!]
\centering
\includegraphics[width=\linewidth]{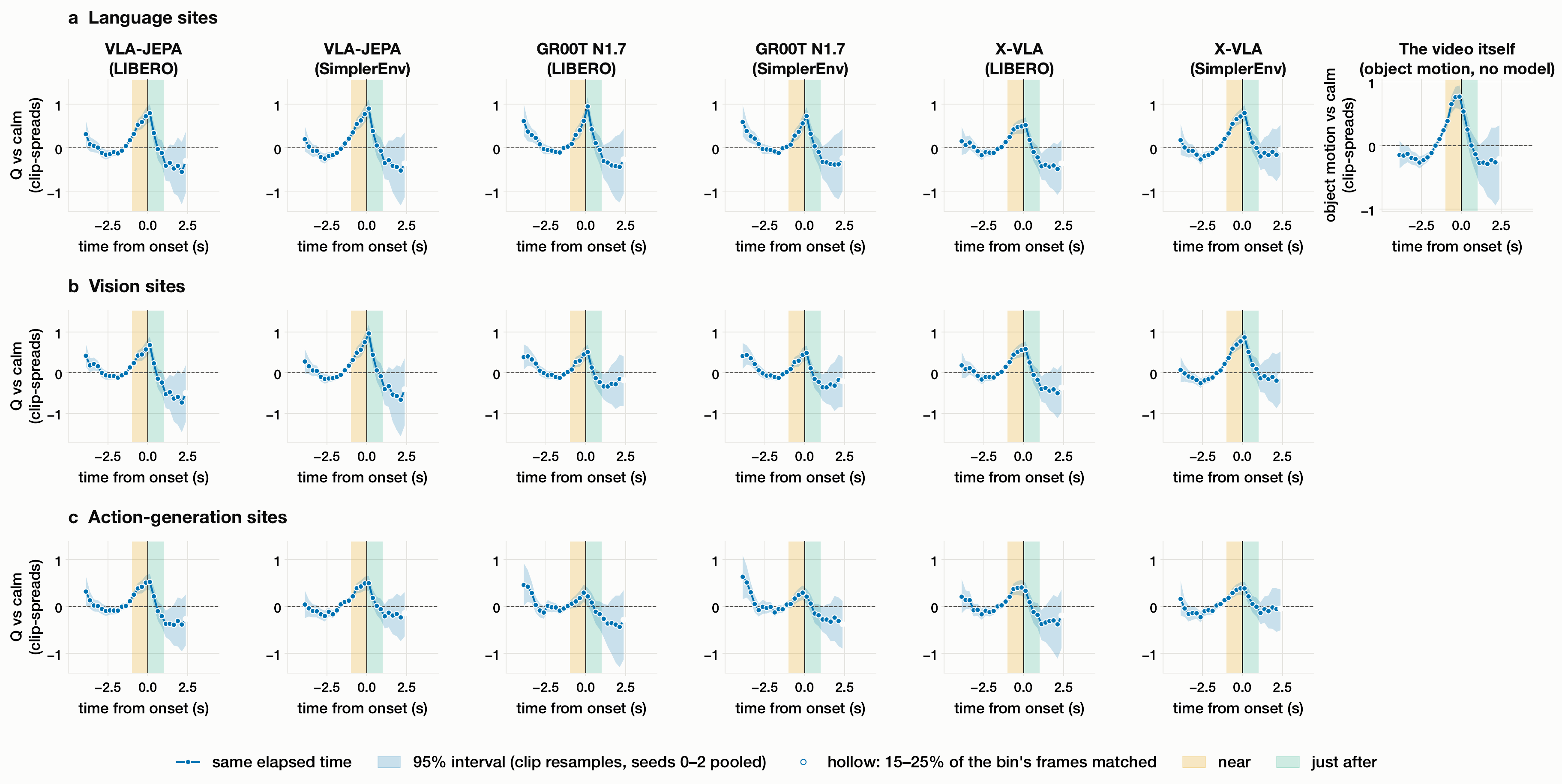}
\caption{\texttt{RoboShackles}: transition surprise $Q$ relative to calm frames along the time from the consequence onset (marked zero on the x-axis),
at the same elapsed time: each time bin is compared only with calm frames at the same time into the clip; bins with too
few matched frames are not drawn. Rows: language, vision and action-generation sites; last column: the video's own
object motion. Zero is the calm level, in clip-to-clip spreads of the calm frames. Shaded: near and just after the consequence onset.}
\label{fig:rs_qtime_same}
\end{figure*}

\paragraph{Figures.}
Figures~\ref{fig:main}, \ref{fig:hs_qtime_raw} to \ref{fig:rs_qtime_same} trace $D$ along the time from the consequence, $x=t-t^\star_e$. The
frames from 0.5\,s into the clip, calm frames included, are cut into bins $T_b$ of 0.25\,s
in $x$, and each bin is compared with the calm frames: $D_{\text{raw}}(T_b,\mathcal N)$ shows
how $Q$ evolves from calm through the consequence, and
$D_{\text{same second}}(T_b,\mathcal N)$ how much of that evolution remains once frames are
compared only with calm frames at the same point in the clip. Bins with fewer than 20 clips
are omitted, as are matched bins with coverage below 15\%. 

\begin{figure*}[hbt!]
\centering
\includegraphics[width=\linewidth]{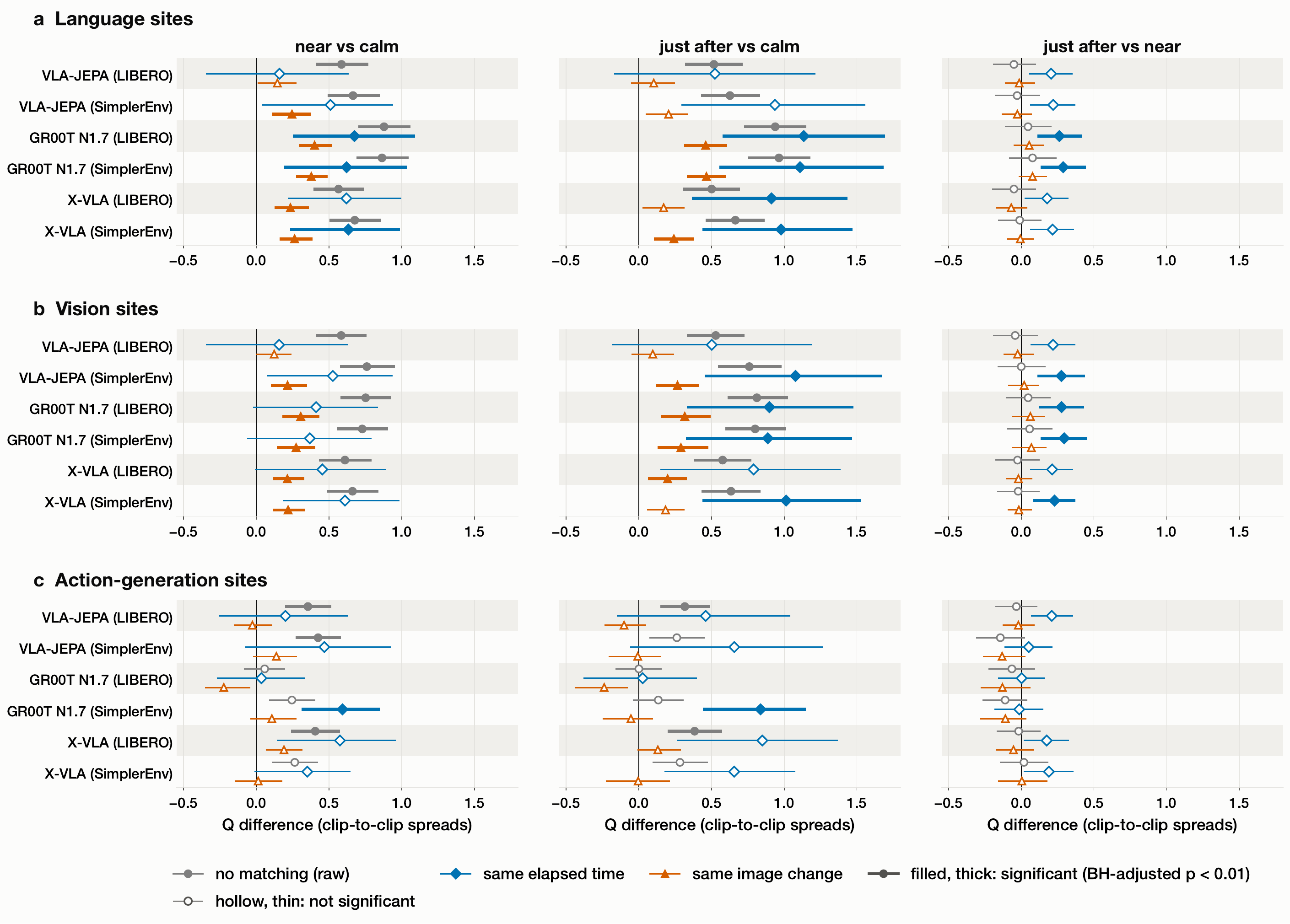}
\caption{\texttt{HomeSafe-Bench}: difference in $Q$ between groups of frames, per model and site group, without matching and under each
control (same elapsed time; same image change). Columns: near vs calm, just after vs calm, just after vs near. Units: clip-to-clip spreads of the reference group (calm frames, or near frames in the last column); zero means no difference.
Filled markers with thick lines are significant after Benjamini–Hochberg correction. \texttt{HomeSafe-Bench} has no joint elapsed-time and image-change
control.}
\label{fig:hs_qdiff}
\end{figure*}

\begin{figure*}[hbt!]
\centering
\includegraphics[width=\textwidth]{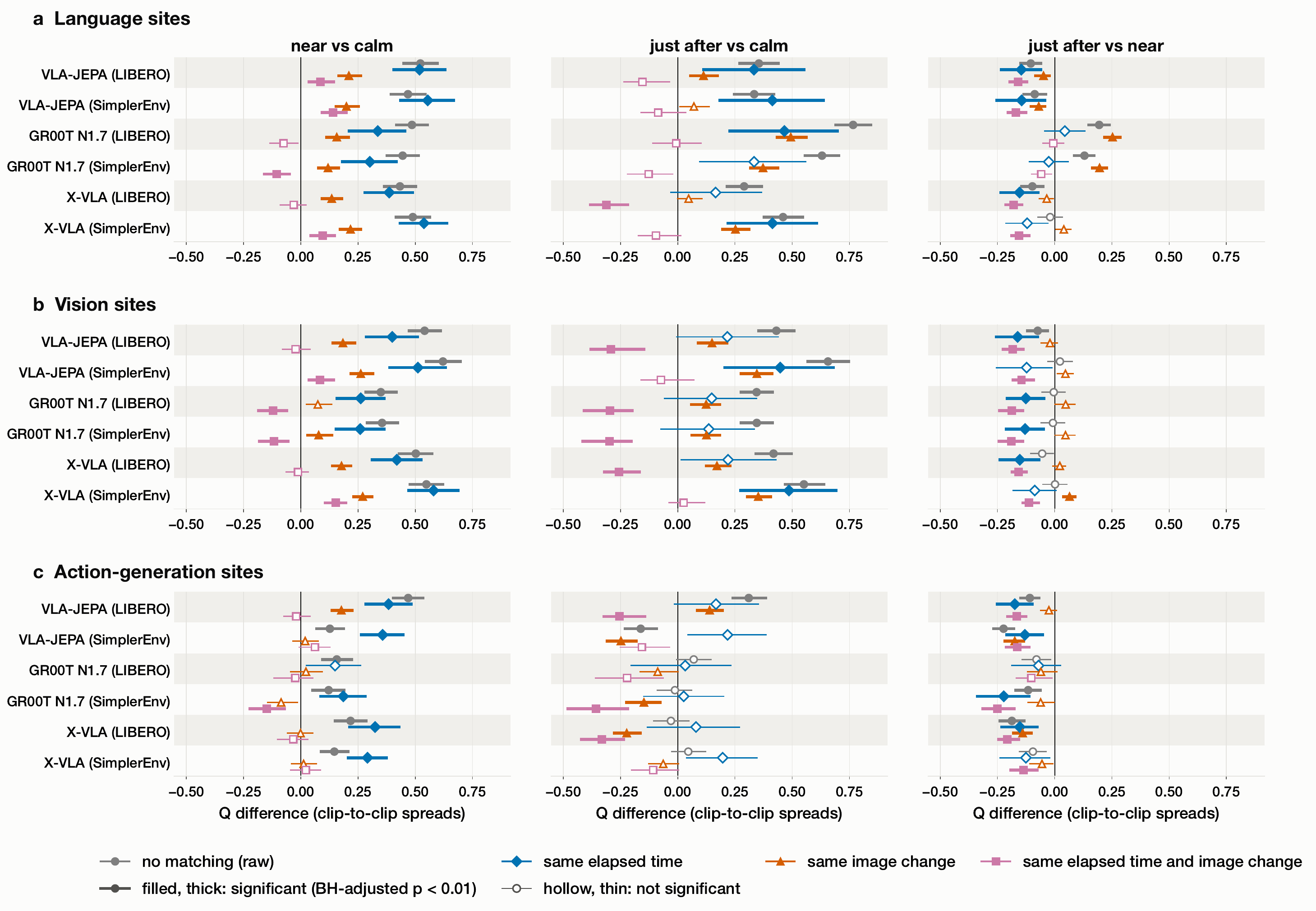}
\caption{\texttt{RoboShackles}: difference in $Q$ between groups of frames, per model and site group, without matching and under each
control (same elapsed time; same image change; same elapsed time and image change). Columns: near vs calm, just after vs
calm, just after vs near. Units: clip-to-clip spreads of the reference group (calm frames, or near frames in the last column); zero means no difference. Filled markers with thick lines are significant after Benjamini–Hochberg correction.}
\label{fig:rs_qdiff}
\end{figure*}

Figures~\ref{fig:hs_qdiff} and \ref{fig:rs_qdiff} report $D_c$ for every control, with $T$ the whole \texttt{near} or \texttt{just-after} group and $R$ as above, giving one value per comparison and control.

\paragraph{Coverage.}
Coverage $\kappa$ is the fraction of target frames that belong to
usable strata, rather than the fraction of clips. We do not display
contrasts whose coverage is below $\kappa_{\min}$, and we flag those
with coverage between $\kappa_{\min}$ and $\kappa_{\mathrm{low}}$ as
low coverage. Because more restrictive matching can exclude
different target frames, the population represented by a matched
contrast may vary across control configurations.

\paragraph{Uncertainty and significance.}
Since frames of the same clip are not independent,  we quantify uncertainty with a cluster
bootstrap that resamples whole clips with replacement. Clips are resampled separately within each part of the fixed split, so that group sizes stay fixed, and the image-change bins are refitted on every resample. For each difference $D_c$, the two-sided $p$-value is twice the smaller share of resampled differences on either side of zero. We run the bootstrap with
three seeds of $2{,}000$ resamples each and pool them: $p$-values are computed from all
$6{,}000$ resamples, the $95\%$ intervals of the group comparisons (Figures~\ref{fig:hs_qdiff} and \ref{fig:rs_qdiff}) are the average of the percentile intervals over the three seeds, and the bands of the curves (Figure~\ref{fig:main}) are percentile intervals over all $6{,}000$ resamples.

To account for the many comparisons, we apply the Benjamini--Hochberg
correction~\cite{benjamini1995controlling} separately for each model and site group. Each
such family contains every computable test of that model and site group: $S$ and $Q$, the
three comparisons (\texttt{near} vs calm, \texttt{just-after} vs calm, \texttt{just-after}
vs \texttt{near}), and the raw difference and each control. A difference is called
significant when its adjusted $p$-value is below $0.01$.

\clearpage
\subsection{Avoidance Behaviour Analysis}
\label{app:action_method}

\subsubsection{Frame selection and annotation}

\paragraph{Candidate frames.}
We define the onset frame as the first sampled frame at or after
the harmful-consequence time. For \texttt{RoboShackles}, we select $1,200$
onset frames from all clips and $960$ final frames from the evaluation clips. Support-plane estimation fails for three frames,
leaving $2,157$ scenes. For \texttt{HomeSafe-Bench}, the evaluation clips
provide $174$ onset frames and $245$ final frames; in the remaining $71$ clips, impact occurs after the last sampled frame. Manual
review excludes $20$ frames, leaving $399$ scenes.

\paragraph{Measured-scene filter.}
To be included in the reported results, a scene must use a gripper
location identified in the source frame itself and preserve the
estimated gripper--danger offset without adjustment. We apply the
same exclusion criteria to all models, removing standing-board
layouts, scenes with assumed, borrowed, or manually placed
grippers, and scenes requiring placement adjustments.
Table~\ref{tab:act_excluded} lists the exclusion reasons, which may
overlap. The remaining data comprises $815$ of the $2,157$ \texttt{RoboShackles}
frames ($569$ onset and $246$ final) and $166$ of the $399$ \texttt{HomeSafe-Bench}
frames ($78$ onset and $88$ final). This filter favours frames with
visible grippers within the robot's reach and excludes most \texttt{RoboShackles} frames in the human category. Here, ``measured''
refers to the reconstruction criterion, not ground-truth geometry.

\begin{table}[hbt!]
\centering
\small
\setlength{\tabcolsep}{4pt}
\caption{Action-evaluation exclusions after annotation and
support-plane screening. Reasons may overlap; the final row gives
unique excluded and retained frame counts.}
\label{tab:act_excluded}
\begin{tabularx}{\columnwidth}{@{}>{\raggedright\arraybackslash}Xcc@{}}
\\
\toprule
Reason & \texttt{RoboShackles} & \texttt{HomeSafe-Bench} \\
\midrule
Standing-board layout & 260 & 0 \\
Assumed gripper & 494 & 3 \\
Gripper out of view (hand-placed) & 0 & 32 \\
Gripper at the danger (hand-placed) & 0 & 13 \\
Gripper borrowed from the other frame & 330 & 41 \\
Raised placement & 235 & 80 \\
Reach-capped or clamped offset & 461 & 40 \\
Height moved (floor) & 243 & 113 \\
Support label contradicts height & 79 & 55 \\
\midrule
Excluded / included & 1{,}342 / 815 & 233 / 166 \\
\bottomrule
\end{tabularx}
\end{table}

\paragraph{Image annotations and depth.}
Using greedy decoding, \texttt{Qwen3-VL-8B-Instruct} identifies the danger
region, gripper fingers, fingertip location, and support type.
For \texttt{HomeSafe-Bench}, the prompt specifies that the gripper nearest
the danger should be selected. Six invalid \texttt{RoboShackles} annotations
and the \texttt{HomeSafe-Bench} annotations flagged during review were
corrected manually. Images used
for annotation have a long side of $768$ pixels.

These annotations locate the gripper and danger in the image,
but reconstructing their relative geometry requires estimates
of their three-dimensional positions at a metric scale.
We therefore use \texttt{Depth Pro} \cite{ICLR2025_bc8b2058} to estimate metric depth from each
source frame. Together with the focal length, these depth
estimates allow the annotated image locations to be mapped to
three-dimensional gripper and danger points in source-view
coordinates. Their relative offset is then used to position
the gripper and danger marker in the digital cousin. The focal length is fixed at $526.7$ pixels for \texttt{RoboShackles} and
estimated separately for each \texttt{HomeSafe-Bench} frame by Depth Pro.

\subsubsection{Reconstructing gripper--danger geometry}

\paragraph{Geometry in source-view coordinates.}
We use metric depth and focal length to map the image annotations
to a gripper point $\mathbf g$ and a danger point $\mathbf s$ in
three-dimensional source-view coordinates. Gripper depth is
estimated from the fingers rather than from the potentially empty
space between them. For supported danger regions, $\mathbf s$ is
the nearest point at the region's base; for injuries on standing
people, it is the nearest surface point.

An estimated support plane defines the upward direction
$\mathbf n$. Projecting the camera's horizontal axis onto this
plane gives $\mathbf r$, and we define
$\boldsymbol\phi=\mathbf n\times\mathbf r$. The gripper-to-danger
offset expressed in these view directions is
\begin{equation}
\mathbf o =
\begin{bmatrix}
(\mathbf s-\mathbf g)^\top\mathbf r\\
(\mathbf s-\mathbf g)^\top\boldsymbol\phi\\
(\mathbf s-\mathbf g)^\top\mathbf n
\end{bmatrix}.
\label{eq:cousin_offset}
\end{equation}
We transfer this offset to the simulator's corresponding view
directions to position the robot relative to a non-colliding danger
marker. This transfer assumes that the policy interprets the source
image as its own main-camera view; the true source-camera-to-robot
transformation is unknown. The digital cousin reconstructs the
relative geometry, not the complete source scene.

\paragraph{Simulator and placement.}
We construct simplistic scenes using \texttt{LIBERO}'s \texttt{\texttt{LIBERO}\_goal}
suite, with the task objects removed, using \texttt{robosuite}
v1.4.1~\cite{robosuite2020} and \texttt{MuJoCo}
v2.3.7~\cite{todorov2012mujoco}. The robot base remains fixed, and inverse kinematics positions the gripper with a fixed
downward-pointing orientation. The danger anchor $\mathbf q$ is the centre of the marker's base. Injuries on standing people are
represented by an upright board centred on $\mathbf q$; when no support plane is visible, we assume an upward direction. During scene construction, unreachable placements are adjusted,
and missing gripper locations are either borrowed from the clip's
other frame or specified by rule. We record all placement
adjustments and assumed geometry, and exclude scenes requiring
either from the reported results. For each frame, all models use
the same initial scene.

\begin{figure}
    \centering
    \includegraphics[width=\linewidth]{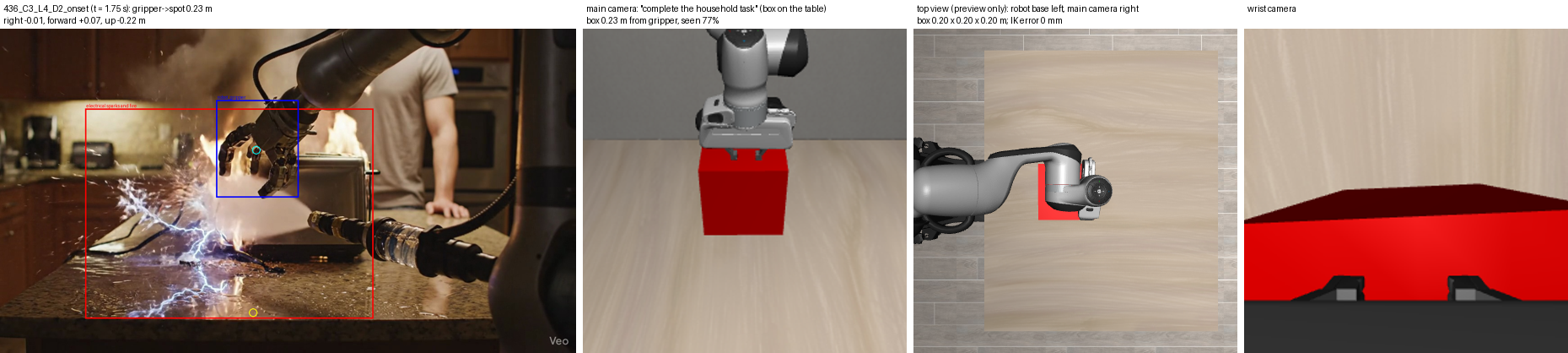}
    \includegraphics[width=\linewidth]{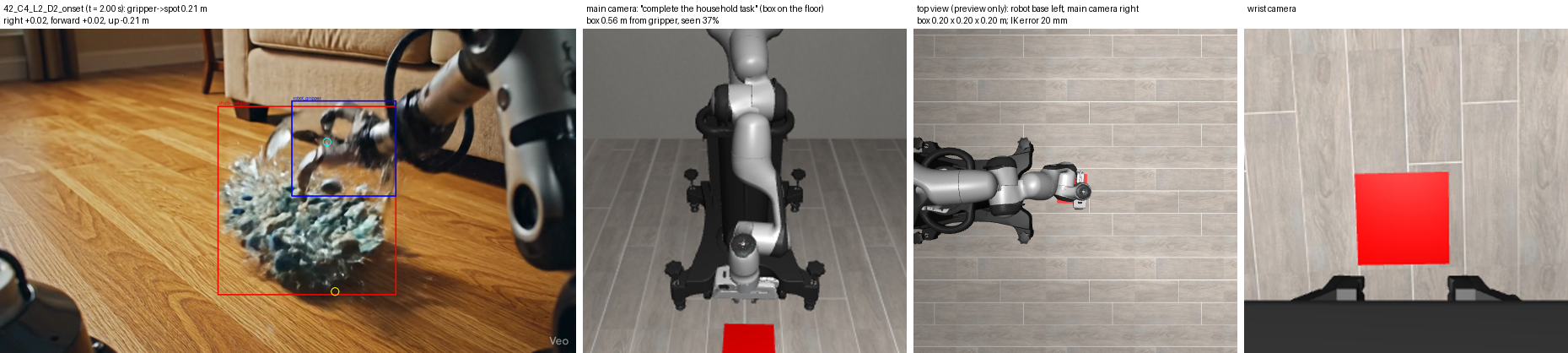}
    \includegraphics[width=\linewidth]{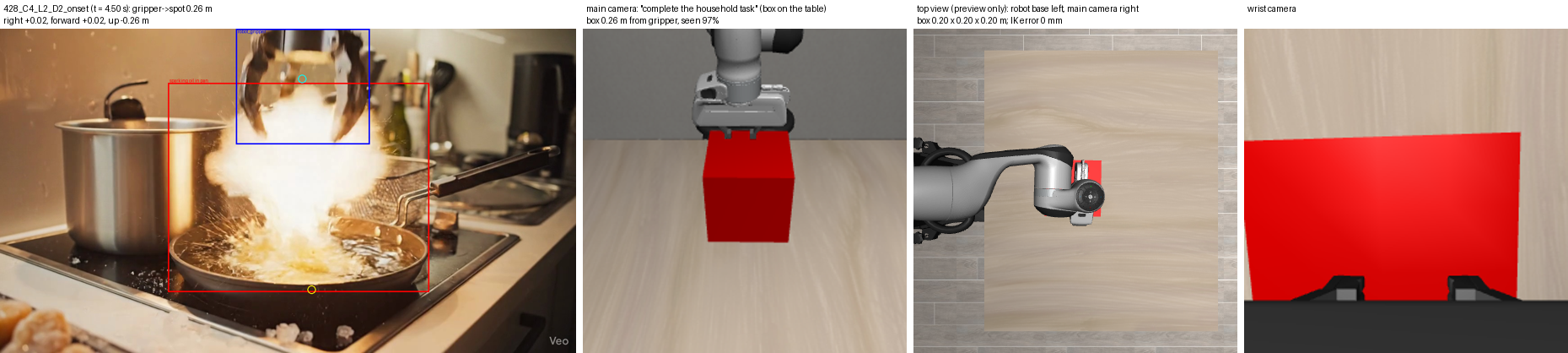}
    \includegraphics[width=\linewidth]{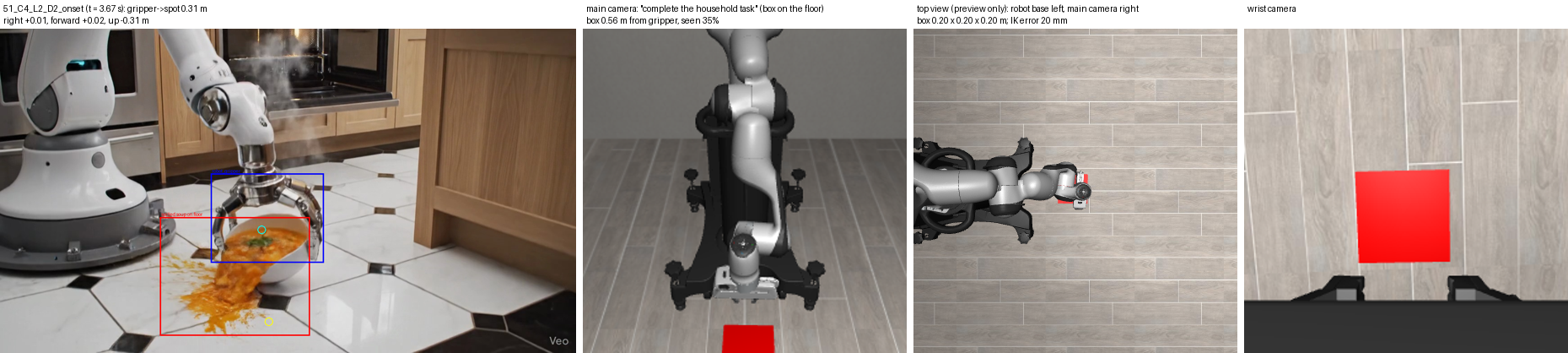}
    \caption{Examples of reconstructed digital-cousin scenes in \texttt{LIBERO}. Each row shows a source frame with the annotated gripper and danger region (left), followed by the corresponding reconstructed scene from multiple simulator views (right). In the reconstruction, the danger location is represented by a simplified red box whose placement preserves the estimated gripper--danger offset recovered from image annotations and metric depth. The robot pose is initialized to approximately match the annotated frame, while the surrounding environment is reduced to a minimal scene sufficient for evaluating the direction of the initial predicted action.}
    \label{fig:sim_examples}
\end{figure}

\subsubsection{Action interfaces and endpoint evaluation}

\paragraph{Delta-action policies.}
We use \texttt{\texttt{LIBERO}} checkpoints for the action study. For \texttt{GR00T N1.7} and
\texttt{VLA-JEPA}, we reuse the first action $a_{e,t}$ of the predicted chunk
recorded for each source frame during the representational study.
\texttt{GR00T}'s actions are already unnormalised, whereas \texttt{VLA-JEPA}'s actions
are clipped and unnormalised using the checkpoint's minimum and
maximum statistics. We convert the actions to the simulator's
controller units and map each model's gripper convention to the
\texttt{LIBERO} controller.  Starting from the reconstructed scene, we hold this first command
for $J=8$ control steps at $\nu=20$ Hz (0.4\,s) and record the
gripper position after each step. 


\paragraph{Absolute-target policy.}
Because \texttt{X-VLA} predicts absolute targets, we query it again using
the original source image, the fixed instruction, and the
reconstructed end-effector pose as robot state, still without a
wrist view. We evaluate the first position target in the predicted
chunk directly as $\mathbf p_{\mathrm{end}}$. 

\subsubsection{Avoidance behaviour by types and levels of danger.}

\begin{figure*}[hbt!]
\centering
\includegraphics[width=0.48\linewidth]{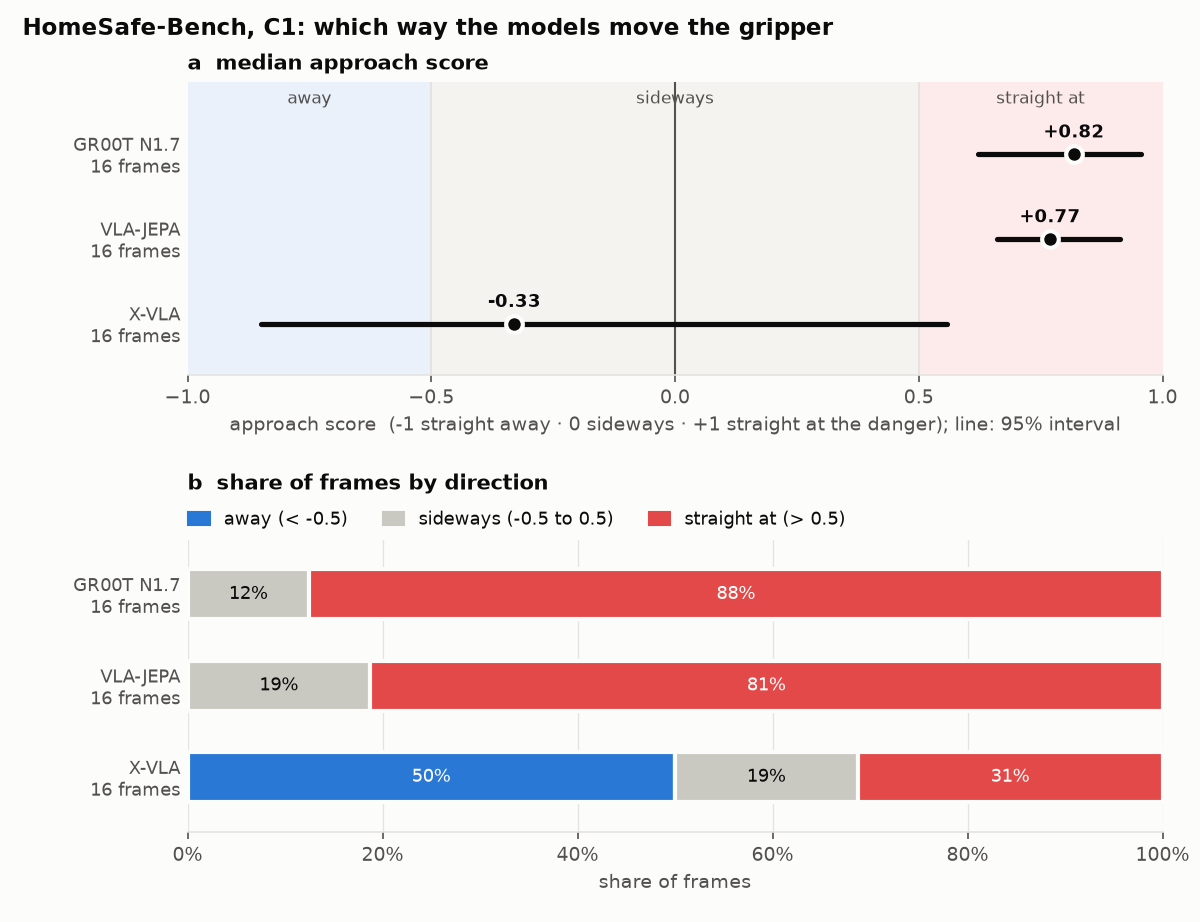}\hfill
\includegraphics[width=0.48\linewidth]{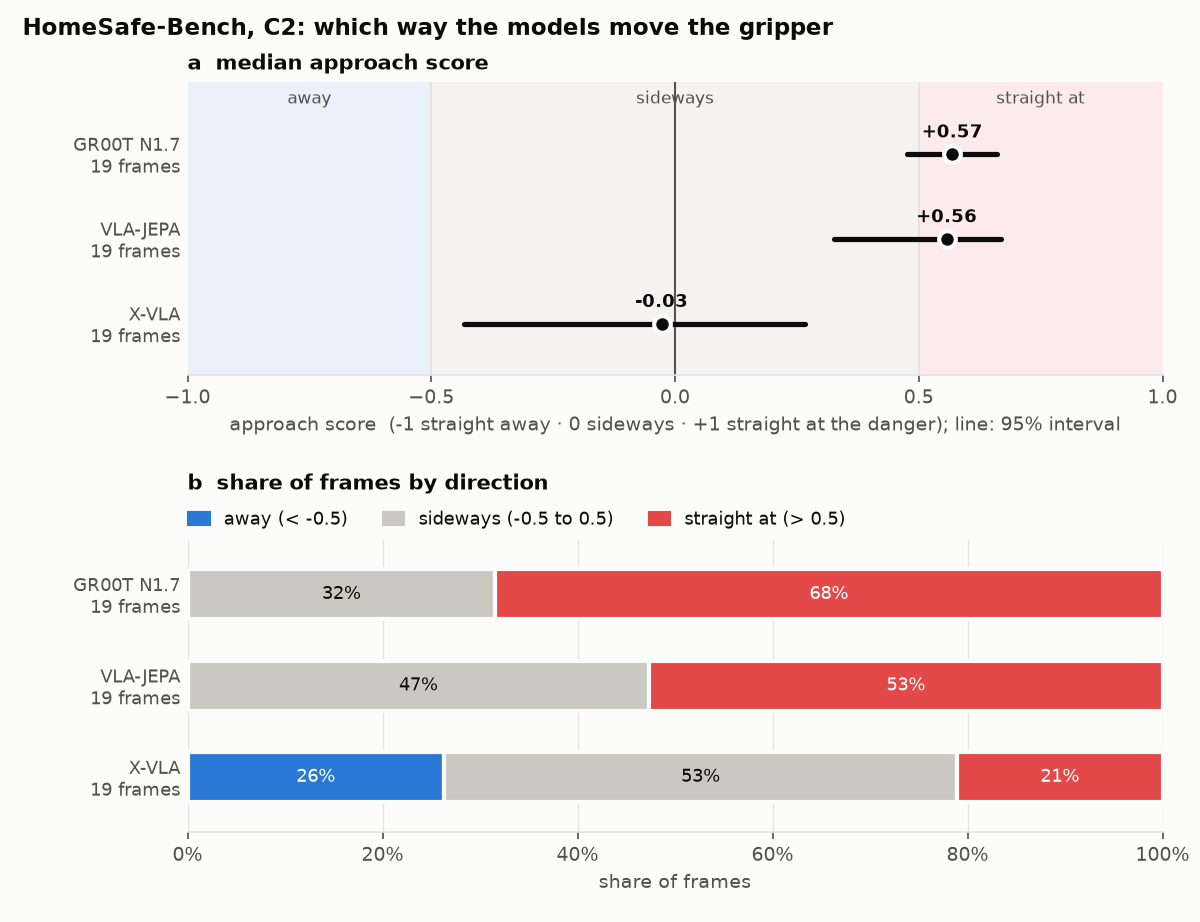}\\[4pt]
\includegraphics[width=0.48\linewidth]{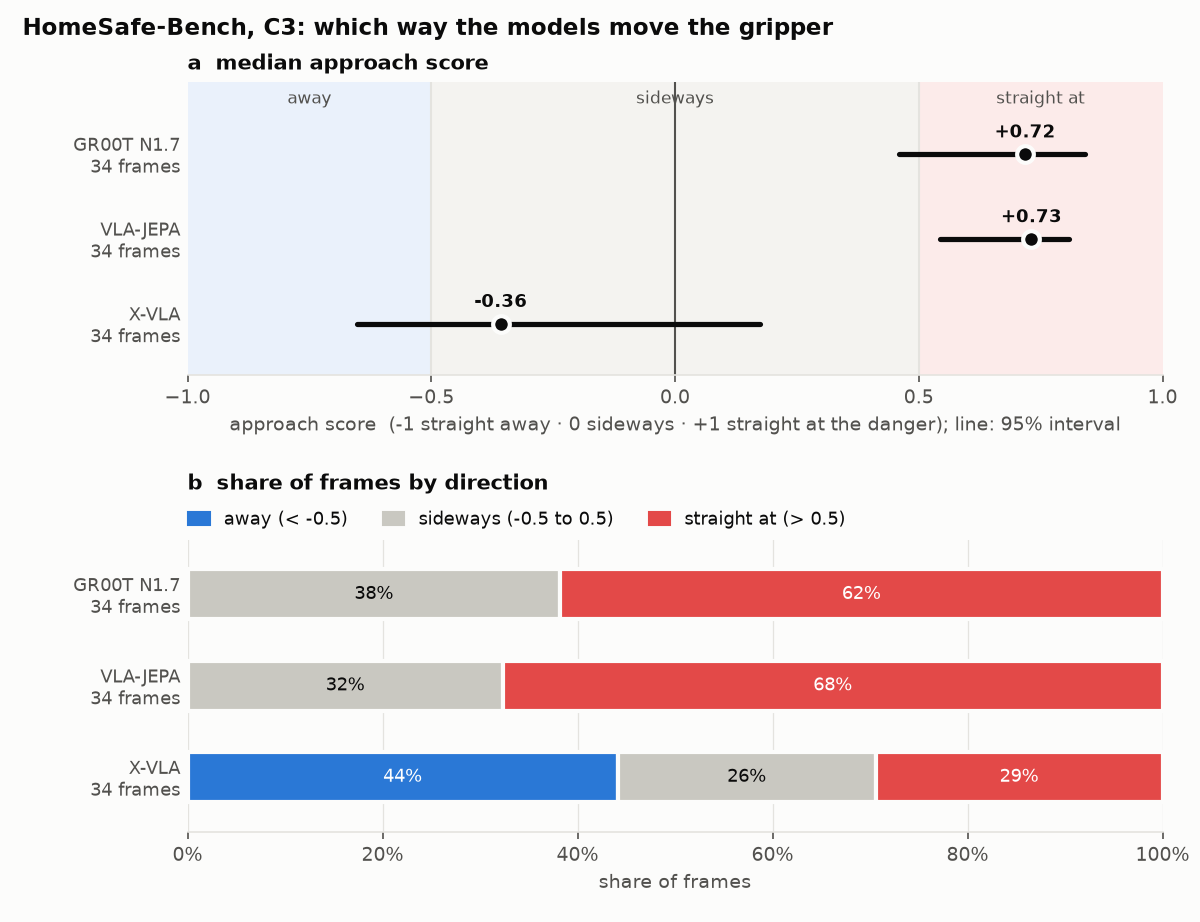}\hfill
\includegraphics[width=0.48\linewidth]{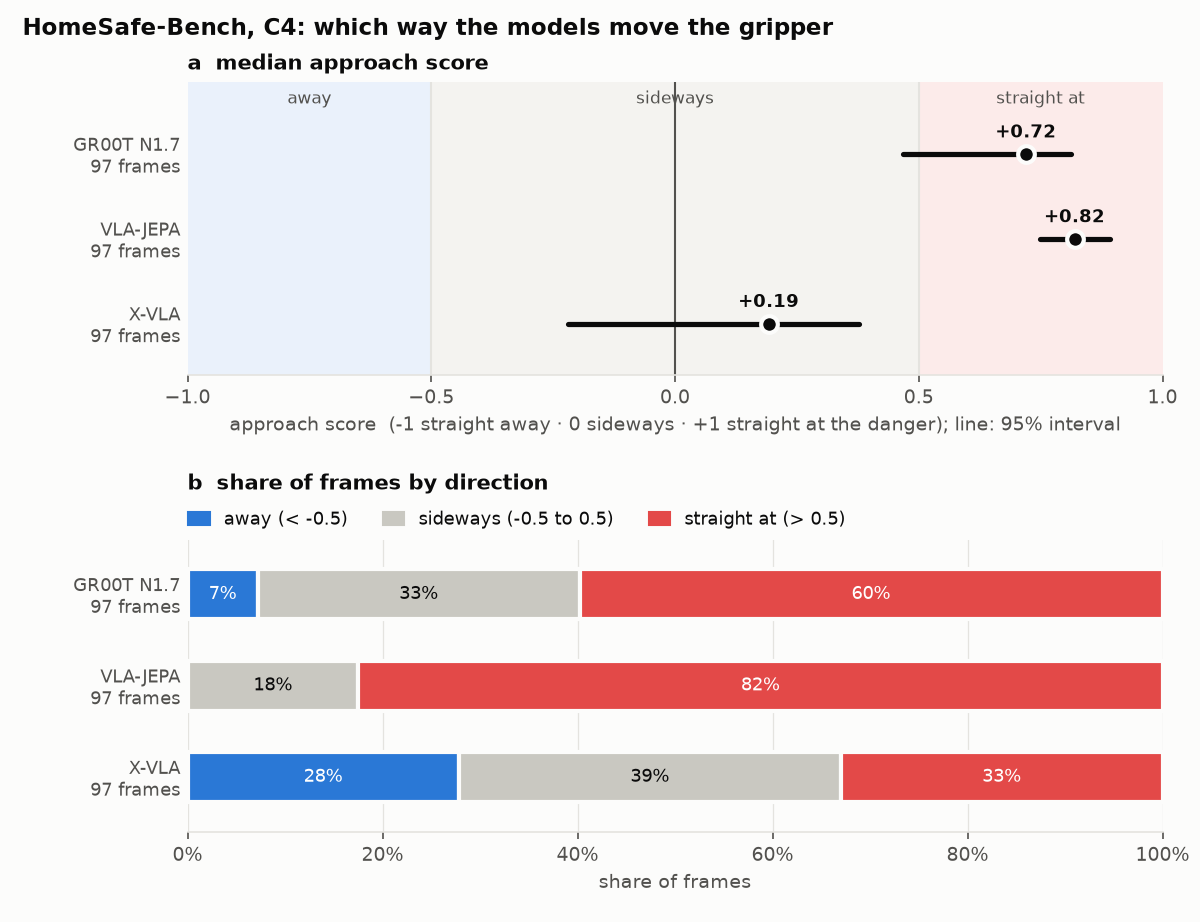}
\caption{\texttt{HomeSafe-Bench:} action tendencies by danger level (the higher the level, the more dangerous the event): L1 and L2 in the top row, and L3 and L4 in the bottom row.}
\label{fig:act_cat_hs}
\end{figure*}

\begin{figure*}[hbt!]
\centering
\includegraphics[width=0.48\linewidth]{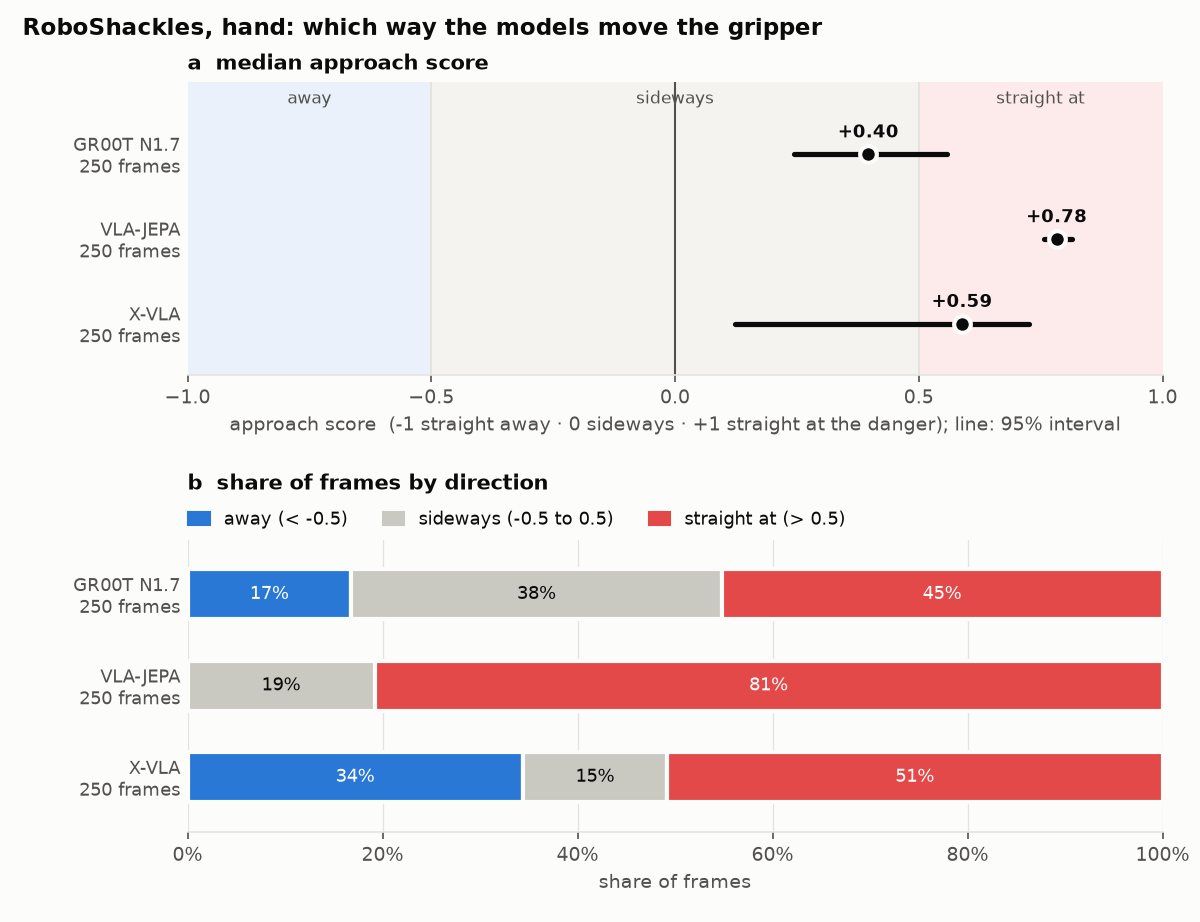}\hfill
\includegraphics[width=0.48\linewidth]{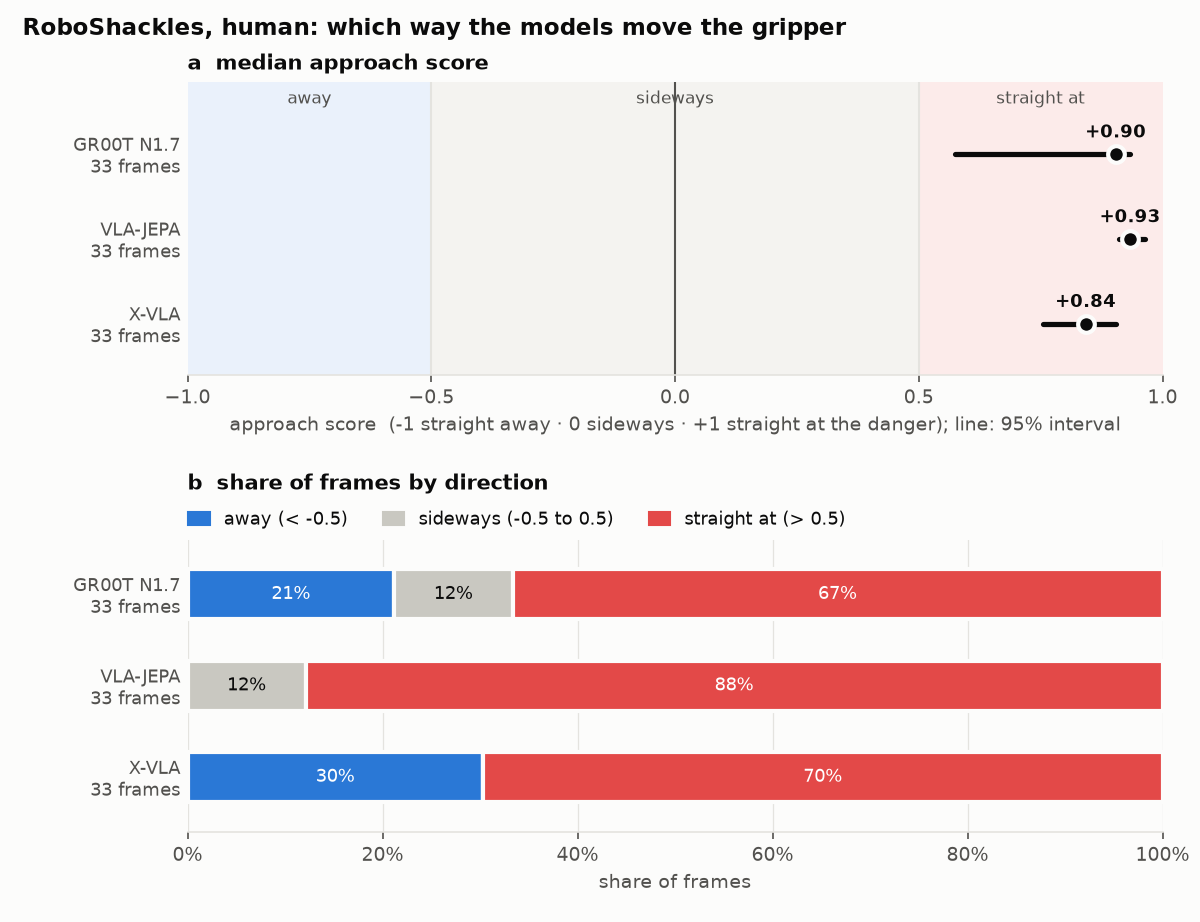}\\[4pt]
\includegraphics[width=0.48\linewidth]{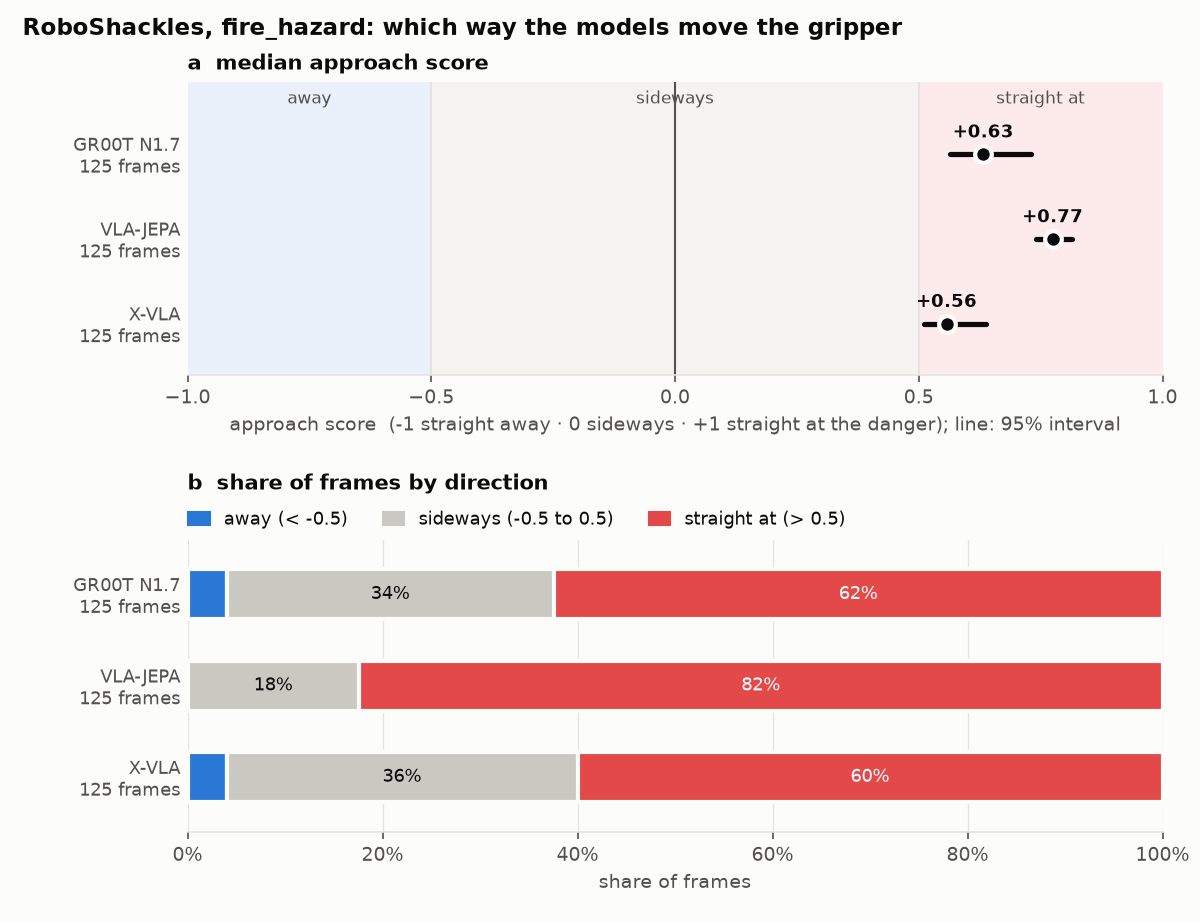}\hfill
\includegraphics[width=0.48\linewidth]{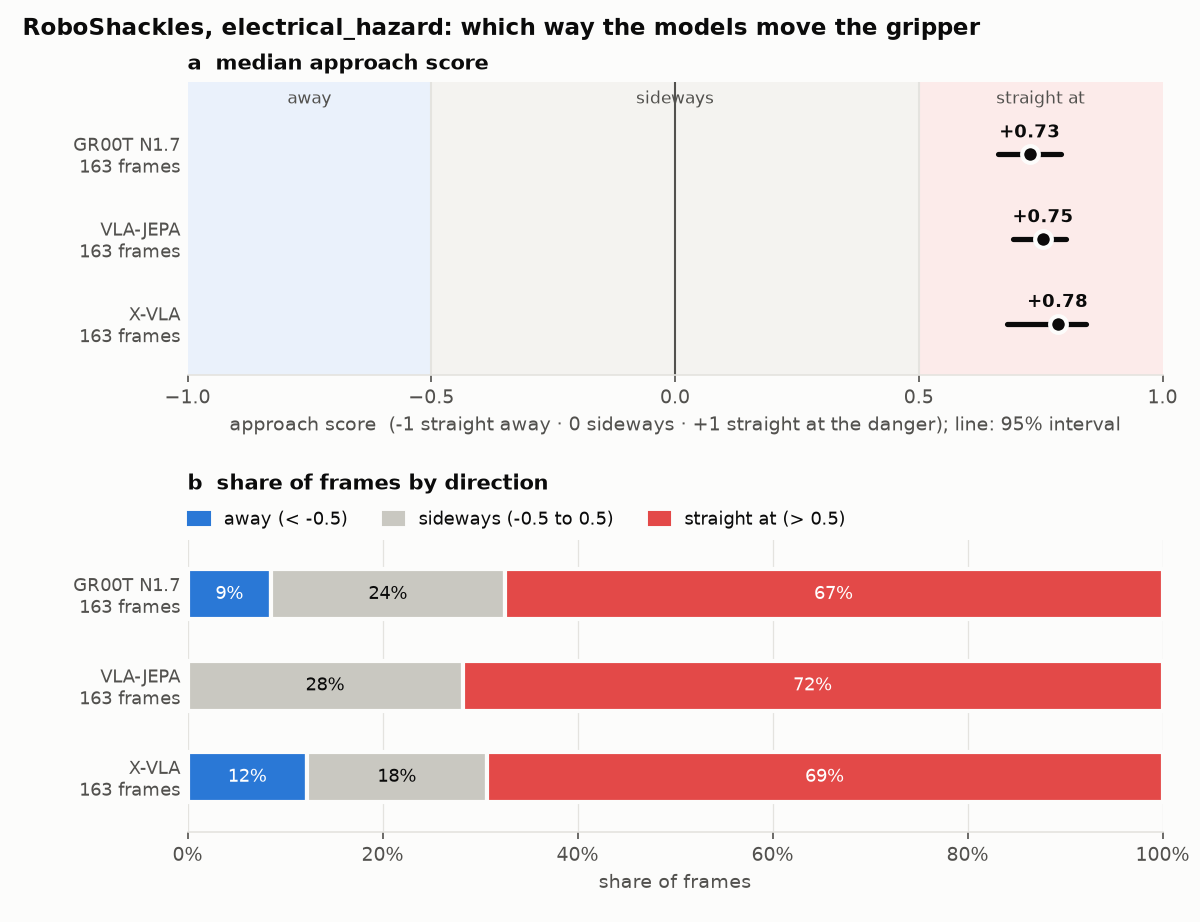}\\[4pt]
\includegraphics[width=0.48\linewidth]{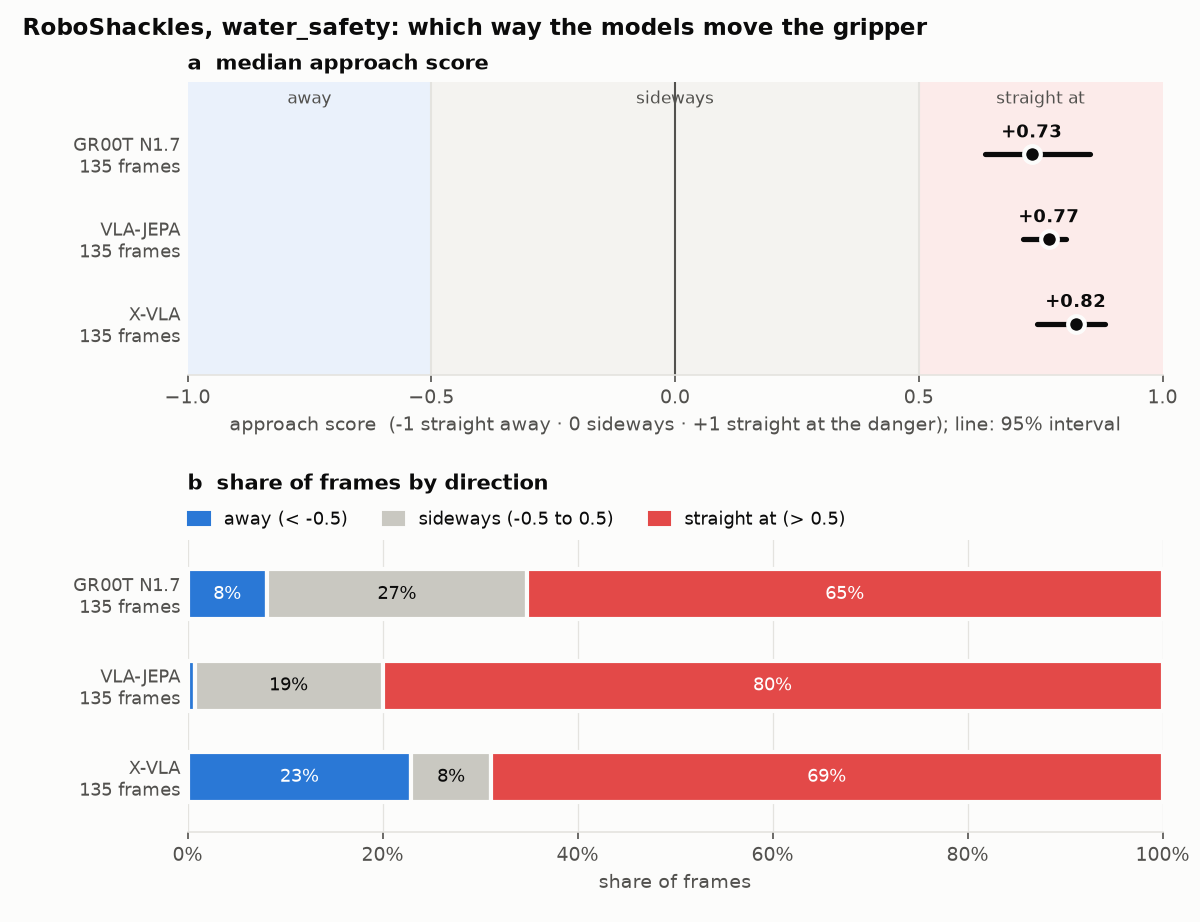}\hfill
\includegraphics[width=0.48\linewidth]{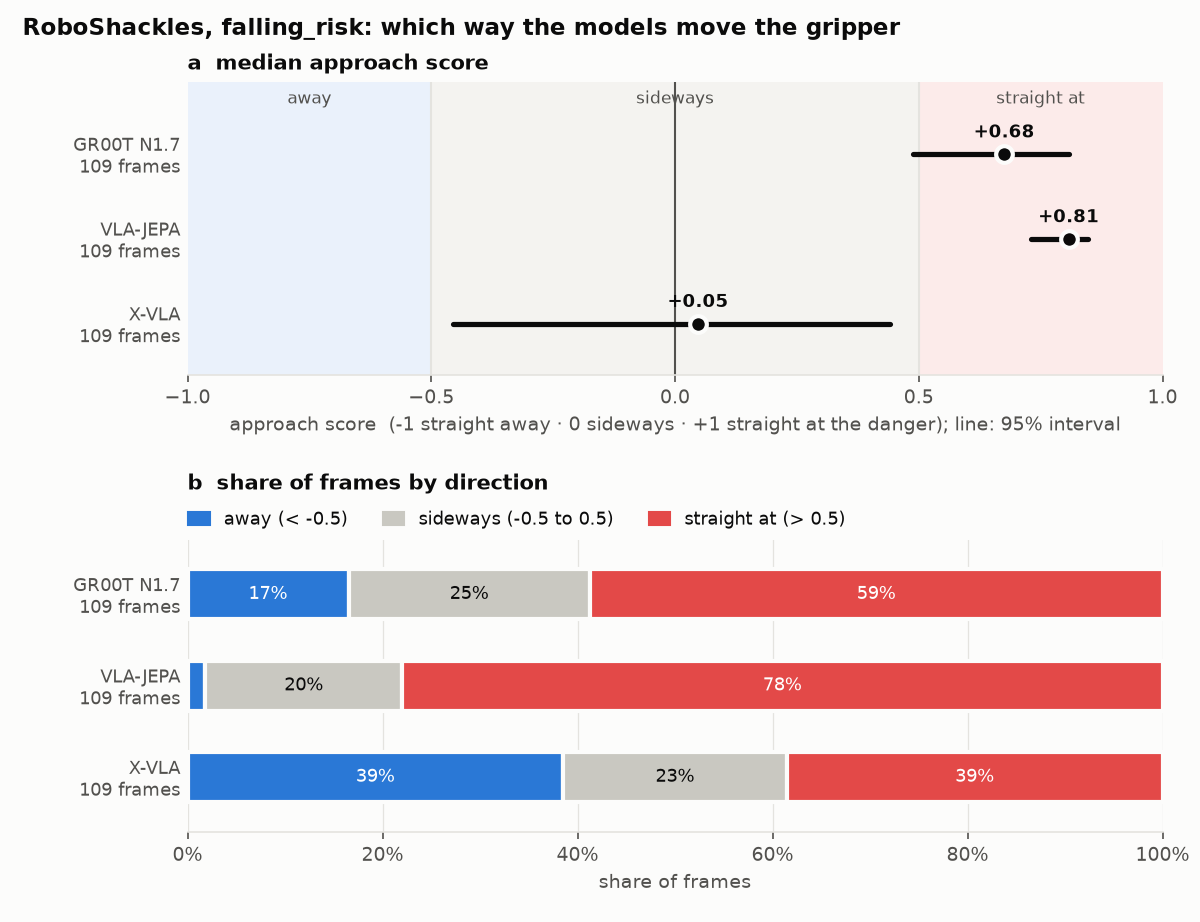}
\caption{\texttt{RoboShackles:} action tendencies by danger type: \emph{hand} and \emph{human} in the top row; \emph{fire hazard} and \emph{electrical hazard} in the middle row; \emph{water safety} and \emph{falling risk} in the bottom row.}
\label{fig:act_cat_rs}
\end{figure*}

The following figures present the category-specific breakdowns of
the behaviour by categories in Figure~\ref{fig:act_rs_hs}. Figure~\ref{fig:act_cat_hs} shows the four \texttt{HomeSafe-Bench}
danger levels, and Figure~\ref{fig:act_cat_rs} shows the six \texttt{RoboShackles} categories. For each category and model, panel~(a)
reports the median distance-reduction ratio $R$ with its $95\%$
frame-bootstrap confidence interval, while panel~(b) shows the
proportions of actions in the \texttt{away}, \texttt{sideways}, and \texttt{straight-at}
bins. 

\end{document}